\documentclass[10pt,twocolumn,letterpaper]{article}

\usepackage{iccv}
\usepackage{times}
\usepackage{epsfig}
\usepackage{graphicx}
\usepackage{amsmath}
\usepackage{amssymb}

\usepackage{booktabs}
\usepackage{dsfont}

\usepackage{multirow} 
\usepackage{multicol}
\usepackage{adjustbox}
\usepackage{subfigure}
\usepackage{pifont}
\usepackage{color}
\usepackage{colortbl}
\usepackage{xcolor}
\usepackage{array}
\definecolor{mygray}{gray}{.9}

\usepackage[pagebackref=true,breaklinks=true,letterpaper=true,colorlinks,bookmarks=false]{hyperref}

 \iccvfinalcopy 

\def\iccvPaperID{3465} 

\ificcvfinal\fi

\begin{document}

\title{Learning Gabor Texture Features for Fine-Grained Recognition}

\author{
    Lanyun Zhu\textsuperscript{\rm1} ~~~
    Tianrun Chen\textsuperscript{\rm2}~~~
    Jianxiong Yin\textsuperscript{\rm3} ~~~
    Simon See\textsuperscript{\rm3} ~~~ 
    Jun Liu\textsuperscript{\rm1}\thanks{Corresponding Author} ~~~ \\
    Singapore University of Technology and Design \textsuperscript{\rm1} ~~~
    Zhejiang University \textsuperscript{\rm2} ~~~
    NVIDIA AI Tech Centre \textsuperscript{\rm3}\\
    {\tt\small lanyun\_zhu@mymail.sutd.edu.sg} ~~~
    {\tt\small tianrun.chen@zju.edu.cn} \\
    {\tt\small \{jianxiongy, ssee\}@nvidia.com} ~~~
    {\tt\small jun\_liu@sutd.edu.sg}\\
}

\maketitle
\ificcvfinal\thispagestyle{empty}\fi

\begin{abstract}
\vspace{-0.7\baselineskip}

Extracting and using class-discriminative features is critical for fine-grained recognition. Existing works have demonstrated the possibility of applying deep CNNs to exploit features that distinguish similar classes. However, CNNs suffer from problems including frequency bias and loss of detailed local information, which restricts the performance of recognizing fine-grained categories. To address the challenge, we propose a novel texture branch as complimentary to the CNN branch for feature extraction. We innovatively utilize Gabor filters as a powerful extractor to exploit texture features, motivated by the capability of Gabor filters in effectively capturing multi-frequency features and detailed local information. We implement several designs to enhance the effectiveness of Gabor filters, including imposing constraints on parameter values and developing a learning method to determine the optimal parameters. Moreover, we introduce a statistical feature extractor to utilize informative statistical information from the signals captured by Gabor filters, and a gate selection mechanism to enable efficient computation by only considering qualified regions as input for texture extraction. Through the integration of features from the Gabor-filter-based texture branch and CNN-based semantic branch, we achieve comprehensive information extraction. We demonstrate the efficacy of our method on multiple datasets, including CUB-200-2011, NA-bird, Stanford Dogs, and GTOS-mobile. State-of-the-art performance is achieved using our approach.
\vspace{-0.7\baselineskip}

\end{abstract}
\vspace{-0.5\baselineskip}
\section{Introduction}
\vspace{-0.5\baselineskip}
Recognizing fine-grained categories (e.g., different flower types or bird species) is challenging due to the subtle differences in their visual appearance. Recently, deep neural network technology \cite{he2016deep, dosovitskiy2020image, liu2021swin, zhu2023continual, zhang2023painting, fu2022panoptic} has made remarkable advancement, allowing for significant progress in extracting deep class-discriminative features for fine-grained recognition \cite{dubey2018maximum, sun2018multi, wang2020weakly, ji2020attention, chen2019destruction, zheng2019looking}. For example, \cite{zhao2021graph} develops a graph-based module to extract cross-class relationship features.  \cite{gao2016compact, kong2017low} use bilinear pooling to exploit high-order features from different networks. \cite{zheng2019learning, sun2018multi} leverage multi-scale information to enhance feature effectiveness. The success of these methods highlights the crucial role of extracting powerful features in distinguishing fine-grained categories.

To exploit class-discriminative features, previous methods \cite{ sun2018multi, wang2020weakly, ji2020attention, chen2019destruction, zheng2019looking} have primarily relied on convolutional neural networks (CNNs) as the model backbone. CNNs have achieved significant success in many computer vision tasks. However, due to the following limitations, they may not be sufficient for capturing comprehensive information in fine-grained recognition. First, deep CNNs generally extract information with high receptive fields and resolution-reduced feature maps, leading to the loss of local detailed information that is critical for fine-grained recognition. Second, as discussed in \cite{luo2019theory,xu2019frequency,xu2019training2}, due to the regularity of the commonly used activation \cite{xu2019frequency} and loss functions \cite{luo2019theory}, CNNs prioritize learning low-frequency components while ignoring high-frequency components. This bias is detrimental to fine-grained
classification, as the ignored high-frequency information, such as ripples and spots in bird feathers, can be important in distinguishing between similar classes. Considering these limitations of CNNs, we suggest that a more powerful feature extractor is required, which should be able to capture local detailed information and multi-frequency comprehensive features that are lost by CNNs but crucial for fine-grained recognition.

Witnessing the challenge, in this work, we introduce a novel branch that extracts texture features as complimentary to the vanilla CNN features to facilitate fine-grained recognition. We employ Gabor filters as the texture extractor in the novel branch. The Gabor filter is a windowed Fourier transform that combines a sinusoidal signal with a Gaussian wave. It is widely used in image processing for extracting low-level texture features \cite{liu2003independent, oh2018analytic, liu2004gabor, arrospide2013log, tao2007general}. The motivations of using Gabor filters in fine-grained recognition can be illustrated in two-folds. (1) First, Gabor filters are effective in capturing local-detailed information. Previous research \cite{daugman1980two, daugman1985uncertainty} has found that Gabor functions are similar to the receptive field profiles in the mammalian cortical simple cells. This indicates that Gabor filters can perform the similar function as cortical cells to capture effective and locally detailed texture information, which is critical for recognizing confusing samples. (2) Second, Gabor filters can extract sufficient high-frequency information. A Gabor filter contains a parameter that directly controls the frequency of the extracted  information. It is therefore possible to exploit sufficient high-frequency components by constraining the parameters of some filters into a high-value range, overcoming the frequency-bias limitations of traditional CNNs. Benefiting from these advantages, our introduction of Gabor filters enables texture features to be extracted as an effective supplement to the typical CNN features for improving fine-grained recognition.

Utilizing Gabor filters in an effective manner is non-trivial. Previous methods with Gabor filters generally rely on hand-crafted designs for filter parameter setting, which cannot guarantee to be optimal. To overcome this limitation, we propose an approach to automatically learn parameters in a supervised manner, in which a value constraint strategy is used to stabilize the learning and enhance high-frequency information extraction. To extract effective information with Gabor filters, we further propose a statistical feature extractor to capture texture features from the Gabor response maps. We also introduce a gate mechanism that selects a limited number of appropriate regions as inputs to the texture branch, thereby reducing computational costs and avoiding information redundancy. These careful designs allow Gabor filters to benefit fine-grained recognition in an effective and efficient manner.

To the best of our knowledge, our work pioneers to learn texture features using Gabor filters in the task of fine-grained recognition. We perform extensive experiments on four datasets, which demonstrate the effectiveness of our method. Benefiting from the powerful texture features from the Gabor filters, our network can leverage the informative and comprehensive information for recognition, resulting in state-of-the-art (SOTA) performance on all datasets. To summarize, our contributions are as follows. (1) First, we pioneer learning Gabor filters in fine-grained recognition. The texture features extracted from Gabor filters serve as an effective supplement to the vanilla CNN features, significantly improving recognition performance. (2) Second, to enhance model effectiveness and efficiency, we implement several key designs based on Gabor filters, including the parameter learning method, a powerful Gabor feature extractor and a region selection gate. (3) Third, extensive experiments demonstrate the effectiveness of our method with state-of-the-art (SOTA) performance achieved on four datasets.

\vspace{-0.5\baselineskip}
\section{Related Work}
\vspace{-0.5\baselineskip}
\noindent \textbf{Fine-Grained Recognition.} 
The goal of fine-grained recognition is to classify classes that are visually similar. Recent methods for the task can be categorized into two major types: feature-encoding methods \cite{gao2016compact, yu2018hierarchical, kong2017low, dubey2018pairwise, dubey2018maximum, sun2018multi, wang2020weakly, ji2020attention, chen2019destruction, zheng2019looking} and localization methods \cite{huang2016part, wei2018mask, branson2014bird, huang2020interpretable, xiao2015application, ding2019selective, simon2015neural, fu2017look}. Feature-encoding methods emphasize on extracting effective features for the better classification results. For example, some methods \cite{gao2016compact, yu2018hierarchical, kong2017low} use bilinear pooling to obtain informative high-level features. Other works \cite{dubey2018pairwise, dubey2018maximum} use feature constraints to enhance feature effectiveness. \cite{sun2018multi, wang2020weakly, ji2020attention, chen2019destruction} extract multi-level features to get comprehensive information. \cite{zhao2021graph} exploits cross-class relationships to capture class-discriminative features. Localization methods emphasize on selecting the most effective regions to capture features. \cite{huang2016part, wei2018mask, branson2014bird} use extra annotations such as bounding boxes and key points to promote the localization ability of the network.  \cite{huang2020interpretable, xiao2015application, ding2019selective, simon2015neural} select informative regions through activation maps. Different from these works, our method innovatively learns Gabor filter to effectively capture texture information. Additionally, the region selection gate in our method can be end-to-end trained without the need for additional annotation, giving it a unique advantage.

\noindent \textbf{Gabor Filters. }
The Gabor filter is an effective tool for extracting texture features, which makes it widely used in various computer vision applications such as face recognition \cite{liu2003independent, oh2018analytic, liu2004gabor}, vehicle verification \cite{arrospide2013log}, object detection \cite{weber2001quadratic, jain1997object, sun2005road} and gait recognition \cite{tao2007general}. Recently, some works \cite{hu2020gabor, luan2018gabor, yuan2022adaptive, zhao2021fractional} introduce Gabor filters into deep neural networks. However, these approaches have suffered from problems like unstable training, frequency bias and high computation costs, leading to low effectiveness and efficiency. Our method addresses the challenges faced by previous approaches through our careful designs, making Gabor filters to be effectively applied in fine-grained recognition.

\begin{figure*}
    \centering
    \includegraphics[width=1.0\linewidth]{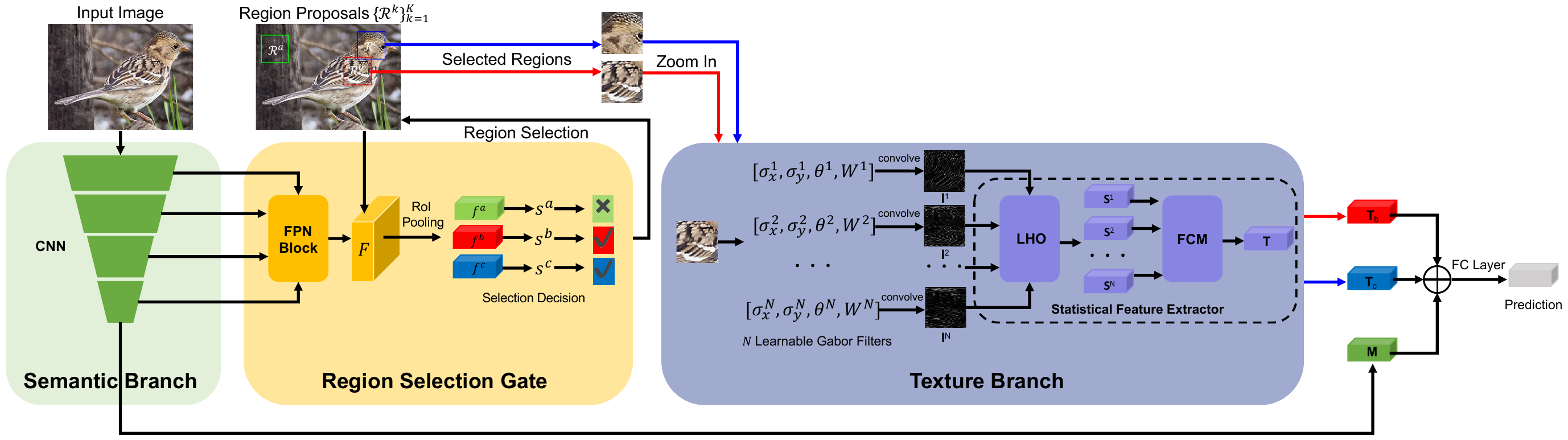}
    \caption{\textbf{Overall structure of our network.} The semantic branch produces a semantic feature $\mathbf{M}$ for the input image. The region selection gate selects impactful regions from a set of pre-defined region proposals. The texture branch generates a texture feature $\mathbf{T}$ for each of the selected regions that is zoomed into the size $S\times S$. Finally, semantic feature $\mathbf{M}$ and texture features $\mathbf{T}$ from all selected regions are added for prediction. Note that, for simplicity of illustration, the figure only presents three region proposals for an image. In actual practice, the region proposals follow the same setting as RCNN \cite{girshick2014rich}, comprising multiple regions with different positions and scales.}
    \label{main_fig}
\end{figure*}

\vspace{-0.5\baselineskip}
\section{Method}
\vspace{-0.5\baselineskip}
\subsection{Overall Structure}
\vspace{-0.5\baselineskip}
Fig. \ref{main_fig} provides an overview of our approach. A semantic branch takes the entire image as input and generates a semantic feature $\mathbf{M}$ through a vanilla CNN. Using intermediate features from the semantic branch, a gate mechanism selects impactful regions from the image. After that, each of the selected regions $i$ is zoomed into the size $S\times S$ and fed into a texture branch to produce a texture feature $\mathbf{T}_{i}$. The texture branch comprises $N$ learnable Gabor filters followed by a statistical feature extractor that performs the following two-step workflow. First, each learnable Gabor filter is convolved with the input region to generate an intensity map $\mathbf{I}$. In this way, for all the $N$ learnable Gabor filters, a set of maps $\{\mathbf{I}^{n}\}_{n=1}^{N}$ are generated. Next, the statistical feature extractor processes each $\mathbf{I}^{n} \in \{\mathbf{I}^{n}\}_{n=1}^{N}$ through a Learnable Histogram Operator (LHO) and then passes the resulting $\{\mathbf{S}^{n}\}_{n=1}^{N}$ through a Filter Correlation Module (FCM) to generate the texture feature $\mathbf{T}$. Finally, the semantic feature $\mathbf{M}$ and texture features $\{\mathbf{T}_{i}\}$ from all selected regions are added and fed into a fully connected layer to obtain the prediction.     

The subsequent sections are arranged as follows. In Sec. \ref{gabor_filer}, we provide an introduction to Gabor filters, which are the key to our method to capture texture information. In Sec. \ref{learn_gabor}, we illustrate the strategy to train our Gabor filters in a stable and effective manner. In Sec. \ref{qco}, we elaborate on the statistical feature extractor that exploits informative statistical information from the output of Gabor filters. In Sec. \ref{region_select}, we introduce the gate mechanism that selects impactful regions into the texture branch for texture extraction. 

\vspace{-0.5\baselineskip}
\subsection{Gabor Filter} 
\vspace{-0.5\baselineskip}
\label{gabor_filer}
The Gabor filter is a widely-used feature extractor in traditional image processing, owing to its biological relevance and powerful texture extraction ability. It has optimal joint localization in both spatial and frequency domains, which makes it possible to be effectively applied in both spatial- and frequency-wise analysis. In the spatial domain, a Gabor Filter is mathematically defined as follows:
\begin{equation}
\label{spatial_gabor}
    g\left(x,y\right) = \frac{1}{2\pi \sigma_{x}\sigma_{y}}\exp\left\{-\frac{1}{2}\left(\frac{\widetilde{x}^{2}}{\sigma_{x}^{2}}+\frac{\widetilde{y}^{2}}{\sigma_{y}^{2}}\right)\right\}\exp[2\pi jW\widetilde{x}].
\end{equation}
\begin{equation}
\left\{
\begin{aligned}
\widetilde{x} & = x\cos \theta + y\sin \theta \\
\widetilde{y} & = -x\sin \theta + y\cos \theta \\
\end{aligned}
\right.
\end{equation}
This representation can be regarded as a windowed Fourier transform, i.e., a Gaussian kernel function modulated by a sinusoidal plane wave.  Specifically, $[\sigma_{x}, \sigma_{y}]$ defines the scale of the Gabor filter, and it can determine the effective size of a pixel's neighborhood where weighted summation occurs. $\theta$ defines the orientation of the filter. $W$ denotes the radial frequency of the sinusoid. These four parameters $\{\sigma_{x}, \sigma_{y}, \theta, W\}$ can fully define a Gabor filter. By altering the parameters, different filters can be generated to capture different signal components.

\subsection{Learnable Gabor Filter Parameters} \label{learn_gabor}
An effective Gabor filter requires the appropriate combination of parameters to match a given task. Previous works \cite{tao2007general, liu2003independent, weber2001quadratic} have generally used a hand-crafted approach for parameter selection, relying on heuristic rules to inform manual parameter settings. However, this kind of hand-crafted approach is highly dependent on the expertise of the individual and cannot guarantee effectiveness for the task at hand. To address this limitation, we propose an alternative method that automatically learns the optimal parameters in a supervised manner, ensuring that the parameter combination can better fit the task. However, learning the parameters presents significant challenges. Firstly, during experimentation, erratic loss fluctuations are observed in the training process, indicating a lack of training stability (see the experimental section for details). Secondly, the features extracted by the trained network exhibit a substantial bias towards low-frequency information. The high-frequency information, on the other hand, occupies only a small fraction of all features. Specifically, only 6\% of the trained Gabor filters have a frequency parameter $W$ higher than 0.5. This tendency is mainly due to the regularity of commonly used activation and loss functions in deep learning (refer to \cite{xu2019frequency, luo2019theory} for details), which is detrimental since the high-frequency information can be vital for fine-grained recognition.

To ensure training stability and filter effectiveness, we propose two solutions to tackle the above problems. To address the first challenge, we propose constraining parameter values into a valid range, which is theoretically derived to avoid potential problems such as frequency aliasing that could hinder training stability. To address the second challenge, we propose a dual-range strategy to further regularize the frequency parameter $W$, ensuring that sufficient high-frequency components can be extracted. 
These solutions enable our Gabor filters to be learned in a stable and effective manner. A detailed illustration of these solutions is presented in the subsequent sections.

\noindent \textbf{Learning Parameters under Constraints. }
We first introduce a novel parameter value constraint to enhance the stability of model training. Our approach is motivated by the analysis that training instability could be attributed to the mismatch between image properties and the ideal application scenario of Gabor filters. Specifically, Gabor filters are mathematically defined for infinite length signals \cite{steiglitz2020digital}; the digital image signals, on the other hand, are finite in both spatial and frequency domains due to the limited length of image width and sampling frequency. As a result, directly applying Gabor filters to finite-length images may cause mathematical deficiency and distortion, leading to frequency aliasing and other issues that can impair model training \cite{sun2005road}. To address this problem, we propose to concentrate most of the Gabor filter energy within the finite signal zone. This ensures that only a small amount of filter energy spills over the finite signal, minimizing the negative effects of using infinite-length-defined filters on finite-length images. Such a constraint could be implemented by limiting the value ranges of filter parameters, which determine the effective zone and energy concentration area of a Gabor filter. Specifically, according to the properties of Gabor filters and images mentioned above, we first derive $[0, \pi]$, $[\frac{5}{2\pi (1-2W)}, \frac{S}{5}]$, $[\frac{5}{2\pi}, \frac{S}{5}]$ and $[0, \frac{2\pi S-25}{4\pi S}]$ to be the valid ranges for parameters $\theta$, $\sigma_{x}$, $\sigma_{y}$ and $W$ respectively (please refer to \textbf{supplementary materials} for a detailed derivation of these valid ranges). Next, we propose a constraint function to ensure each parameter to fall within its valid range. Specifically for a parameter $p$ with lower bound $l_{p}$ and upper bound $u_{p}$, we propose to obtain $p$ through a learnable parameter $\mathbf{p}$ as follows:
\begin{equation}
\label{eq_constrain}
    p = l_{p} + \left(u_{p} - l_{p}\right){\rm Sigmoid}\left(\mathbf{p}\right)=\frac{u_{p} + l_{p}e^{-\mathbf{p}}}{1+e^{-\mathbf{p}}}.
\end{equation}
The parameter $\mathbf{p}$ can be learned to have any value ranging from $-\infty$ to $+\infty$. By applying Eq. \ref{eq_constrain}, parameter $p$ can be guaranteed to fall within its valid range $[l_{p}, u_{p}]$. By doing so, parameter values are constrained to enhance the stability of training.

\noindent \textbf{High Frequency Enhancement. }
To address the frequency bias problem and improve high-frequency information extraction, we further propose a dual-constraint strategy for the frequency parameter $W$. Specifically, we constrain half of Gabor filters to only exploit high-frequency components, making the network to capture sufficient high-frequency information for facilitating fine-grained recognition. This is enabled by setting two value intervals for $W$: $[0, \frac{2\pi S-25}{8\pi S}]$ and $[\frac{2\pi S-25}{8\pi S}, \frac{2\pi S-25}{4\pi S}]$, which are equally divided from the valid range $[0, \frac{2\pi S-25}{4\pi S}]$ of the frequency parameter. Let $N$ be the total number of Gabor filters to be learned. Using the method in Eq. \ref{eq_constrain}, we constrain $W$ of $N/2$ filters to fall between $[0, \frac{2\pi S-25}{8\pi S}]$ and the other $N/2$ filters to fall between $[\frac{2\pi S-25}{8\pi S}, \frac{2\pi S-25}{4\pi S}]$, such that they serve as the low-frequency expert and high-frequency expert respectively. This strategy enhances the extraction ability of high-frequency information, alleviating the frequency bias problem in traditional DNNs where high-frequency features are difficult to extract.

\begin{figure}
    \centering
    \includegraphics[width=1.0\linewidth]{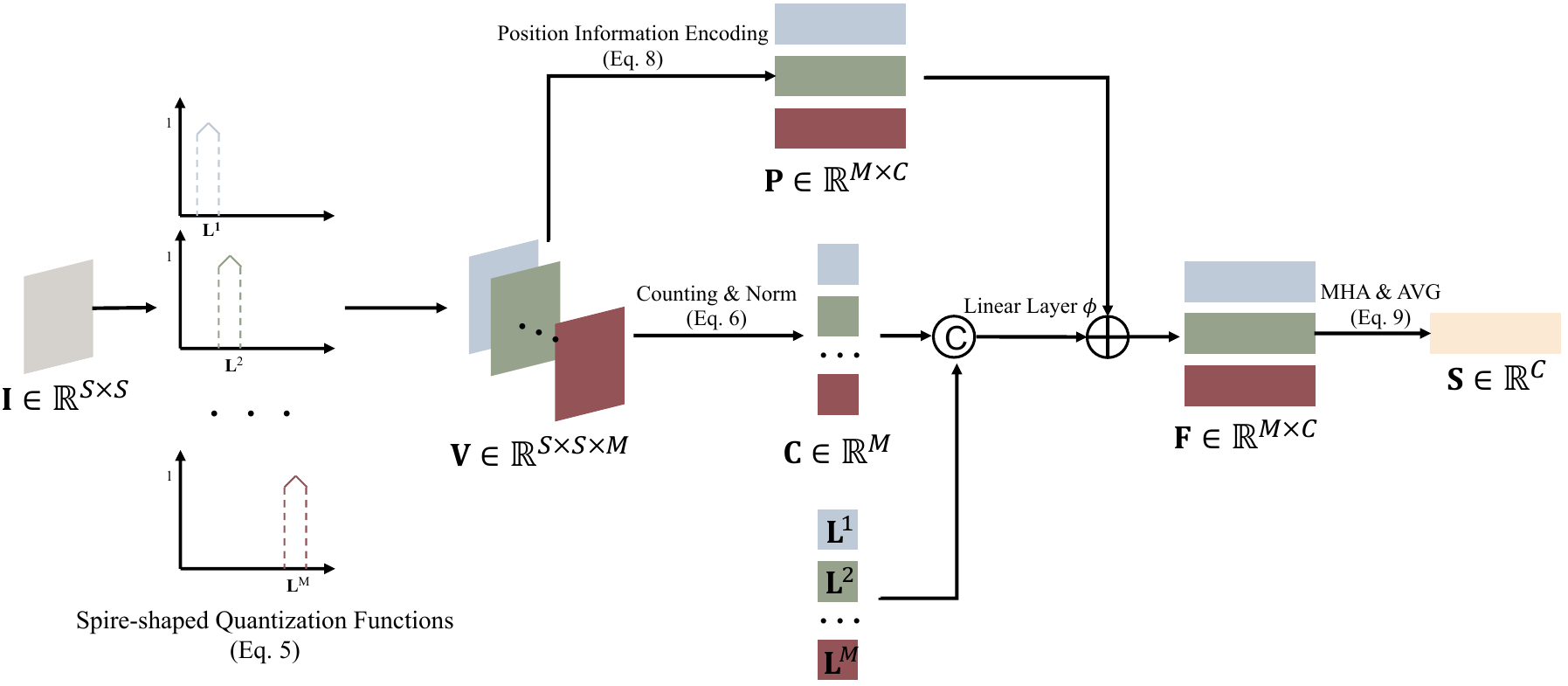}
    \caption{The structure of Learnable Histogram Operator (LHO).}
    \label{qco_fig}
\vspace{-0.7\baselineskip}
\end{figure}

\subsection{Statistical Feature Extractor}\label{qco}
The above method enables Gabor filters to be trained stably and effectively in a deep learning framework. These filters can capture texture signals from an input image. In the next step, we target at extracting effective features from these filters to facilitate fine-grained recognition. From the perspective of digital image processing, texture can be represented by the statistical properties of feature maps \cite{gonzales1987digital}. Based on this principle, in some previous research \cite{tao2007general, clausi2005design, arrospide2013log}, textures were exploited by extracting hand-crafted statistics such as variance, contrast and smoothness from the output of Gabor filters. However, these statistics have limited effectiveness because they are hand-crafted designed. In this work, we step further and take advantage of deep learning to extract deep statistical features that are more comprehensive and effective. Specifically, we propose a novel statistical feature extractor that contains a \textbf{L}earnable \textbf{H}istogram \textbf{O}perator (LHO) followed by a \textbf{F}iler \textbf{C}orrelation \textbf{M}odule (FCM). In the following sections, we provide a detailed illustration of QCO and FCM respectively.

\noindent \textbf{Learnable Histogram Operator.} We first employ a Learnable Histogram Operator (LHO) to exploit statistical information, which is improved from \cite{zhu2021learning}. Fig. \ref{qco_fig} illustrates the structure of LHO. LHO receives the output from each Gabor filter as input and produces a statistical feature $\mathbf{S}\in \mathbb{R}^{C}$, where $C$ refers to the channel number. Let $\mathbf{I} \in \mathbb{R}^{S\times S}$ be the intensity map generated by convolving a Gabor filter with the processed image, where $S$ refers to image size. LHO first defines $M$ levels by equally dividing the minimal and maximal values for intensities of all pixels. Specifically, the $m$-th level $\mathbf{L}^{m}$ is computed as:
\begin{equation}
    \mathbf{L}^{m} = \min\left(\mathbf{I}\right)+\frac{\max\left(\mathbf{I}\right)-\min\left(\mathbf{I}\right)}{M}\cdot m
\end{equation}
 Next, each pixel $\mathbf{I}^{i,j}$ on $\mathbf{I}$ is quantized to a vector $\mathbf{V}^{i,j}\in \mathbb{R}^{M}$. Concretely, let $I$ be the interval between adjacent levels $(I=\mathbf{L}^{m+1}-\mathbf{L}^{m})$, the $m$-th channel $\mathbf{V}^{i,j,m}$ on $\mathbf{V}^{i,j}$ is computed as follows:
\begin{equation}
\mathbf{V}^{i,j,m}=\left\{
\begin{aligned}
    &1-|\mathbf{L}^{m} - \mathbf{I}^{i,j}|\ \ {\rm if}\ \  {|\mathbf{L}^{m} - \mathbf{I}^{i,j}| < \frac{I}{2}}\\
    &0 \qquad \qquad \qquad \ \ {\rm else}
\end{aligned}
\right.
\end{equation}
In this way, $\mathbf{I}$ is quantized to $\mathbf{V}\in \mathbb{R}^{S\times S\times M}$. Note a spire-shaped function is used for quantization instead of a binary one to ensure that the operation is differentiable. After the quantization, the number of pixels that fall into each level is counted and normalized to obtain a counting feature $\mathbf{C}\in \mathbb{R}^{M}$. Specifically, $\mathbf{C}^{m}$ for the $m$-th level is computed as: 

\begin{equation}
    \mathbf{C}^{m} = \frac{\sum_{i=1}^{S}\sum_{j=1}^{S}\mathbf{V}^{i,j,m}}{\sum_{m=1}^{M}\sum_{i=1}^{S}\sum_{j=1}^{S}\mathbf{V}^{i,j,m}}.
\end{equation}
We concatenate $\mathbf{C}^{m}$ with $\mathbf{L}^{m}$, and obtain the feature $\mathbf{F}^{m}$ for the $m$-th quantization level as follows:
\begin{equation}
\label{encode_position}
    \mathbf{F}^{m} = \phi\left({\rm CAT}\left(\mathbf{C}^{m},\mathbf{L}^{m}\right)\right)+ \mathbf{P}^{m},
\end{equation}
where $\phi$ refers to a linear layer. $\mathbf{P}^{m}\in \mathbb{R}^{C}$ encodes the position information for pixels quantized to the $m$-th level, which is computed as the weighted sum of position features from all pixels as follows:
\begin{equation}
    \mathbf{P}^{m} = \sum_{i=1}^{S}\sum_{j=1}^{S}\mathbf{V}^{i,j,m}{\rm PE}\left(i,j\right),
\end{equation}
where ${\rm PE}$ refers to the position encoding operation as in vit \cite{dosovitskiy2020image}. Finally, the statistical texture $\mathbf{S}$ is produced by correlating features from different levels:
\begin{equation}
\label{correlate_level}
    \mathbf{S} = {\rm Avg}\left({\rm MHA}\left(\{\mathbf{F}^{m}\}_{m=1}^{M}\right)\right),
\end{equation}
where ${\rm MHA}$ denotes a multi-head attention and ${\rm Avg}$ refers to the average over all levels. Using LHO has two advantages. Firstly, it can be considered as a way to extract deep features from a differentiable and learnable histogram, which can produce effective and appropriate statistical features that can benefit the task \cite{gonzales1987digital, zhu2021learning}. Secondly, conventional statistical features are typically pixel-relationship-invariant. For example, rearranging pixel positions does not alter the mean and variance values. In contrast, our method is pixel-relationship-aware as it encodes $\mathbf{P}^{m}$ that represents the position information of pixels quantized to each level, which provides additional information that can be useful for fine-grained recognition. These advantages enable the extracted features from LHO to be effective in facilitating fine-grained recognition.

\noindent \textbf{Filter Correlation Module.} The proposed LHO can produce a statistical feature from the output of each Gabor filter, resulting in a set of features $\{\mathbf{S}^{n}\}_{n=1}^{N}$ for all filters. We further propose a Filter Correlation Module (FCM) to aggregate statistical information from all Gabor filters as follows:
\begin{equation}
\label{correlation}
    \mathbf{T} = {\rm Avg}\left({\rm MHA}\left(\left\{\mathbf{S}^{n} + \phi\left(\sigma_{x}^{n}, \sigma_{y}^{n}, \theta^{n}, W^{n}\right)\right\}_{n=1}^{N}\right)\right),
\end{equation}
in which $\mathbf{S}^{n}$ refers to the statistical feature for the $n$-th Gabor filter. $\phi\left(\sigma_{x}^{n}, \sigma_{y}^{n}, \theta^{n}, W^{n}\right)$ uses a linear layer to encode filter parameter information, which can indicate the type of texture extracted by the Gabor filter. Eq. \ref{correlation} exploits the correlations among different texture components from different Gabor filters, generating a powerful feature with long-range correlations that can facilitate fine-grained recognition. 

\subsection{Region Selection Gate} \label{region_select} Through the methods outlined above, we can build a powerful feature extractor that captures texture information using Gabor filters. As a common practice used by previous methods, the extractor can generate features in an image-based \cite{tao2007general} manner or a patch-based manner \cite{sun2005road}. Image-based methods directly feed the whole image into the extractor, which might lead to the blurring of local detailed information. Patch-based methods address the problem by equally dividing the image into several patches and extracting a feature from each patch. However, it is computationally expensive as each image position must be processed by the feature extractor. Intuitively, the texture information in different regions of an image is not equally important. For example, some regions may contain critical textures for recognition, such as feather of the bird, while others may be less informative, consisting of background textures such as sky, road and lawn. Therefore, extracting textures from all positions is redundant and computationally inefficient, which can impede recognition accuracy and slow down algorithmic speed. To address the issue, we propose a novel approach to solely extract textures from informative regions that are automatically selected from a gate mechanism. More specifically, the gate mechanism for region selection involves three steps. Firstly, we introduce a shallow FPN block that takes the intermediate features of the CNN-based semantic branch as inputs and generates a feature map $F$. Please refer to supplementary materials for a detailed description for the structure of this block. Since features from the CNN branch have a high receptive field and contain rich semantic information, $F$ can identify key parts of an image and assist in selecting class-discriminative regions in appropriate positions. Secondly, we introduce a set of regions proposals, denoted as $\{\mathcal{R}^{k}\}_{k=1}^{K}$, which is the same as that in RCNN \cite{girshick2014rich}, comprising regions with different scales and positions. For each region $\mathcal{R}^{k}$, a feature vector $f^{k}$ is generated by performing an RoI pooling on the corresponding region of the feature map $F$. This feature vector is then fed into a linear layer to generate a score $s^{k}$. Lastly, regions that require texture extraction are selected based on their scores. This mechanism enables only the regions that contain class-discriminative information to be selected for texture extraction, thereby reducing computation costs and avoiding information redundancy.

To reduce computation for capturing texture features, we expect `hard' selections instead of the `soft' attention scores to be generated so that only a limited number of regions need texture extraction. However, such a hard selection may prohibit the gradients to be back-propagated to the earlier layers in optimization. To address this issue, we employ the Improved Semantic Hashing \cite{kaiser2018discrete, kaiser2018fast} for region selection. In training, to encourage more decision space to be explored randomly, we add a standard Gaussian noise $\epsilon$ to $s^{k}$, resulting in the noise-injected score $\hat{s}^{k}$. Then two vectors are generated as follows:
\begin{equation}\label{d_c_generation}
    c^{k} = \sigma'\left(\hat{s}^{k}\right);\; d^{k} = \mathds{1}\left(\hat{s}^{k}>0\right),
\end{equation}
where $\sigma'$ refers to the saturating Sigmoid function:
\begin{equation}
    \sigma'\left(\hat{s}^{k}\right)=\max\left(0, \min\left(1, 1.2\sigma\left(\hat{s}^{k}\right)-0.1\right)\right),
\end{equation}
where $\sigma$ denotes the original Sigmoid function. In this way, $d^{k}$ is a binarized discrete feature with value $\{0,1\}$, which is used for making the selection decision. $c^{k}$ is a continuously differentiable vector that can be used to approximate gradients in back-propagation, whose details are presented in supplementary materials.

\noindent \textbf{Discussion with Previous Methods.} Note that the proposed region gate mechanism differs from other key part localization methods used for fine-grained recognition. Unlike existing methods, our method can be trained end-to-end and generate hard decisions without the need for extra annotations. These properties enable our method to be highly effective and practical for region selection. Conversely, existing key part localization techniques have several limitations. For example, some methods \cite{huang2016part, wei2018mask} require extra annotations such as bounding box and key points for joint training; \cite{ge2019weakly} uses an indifferentiable selection process, making the network unable to be trained end-to-end; other methods \cite{ding2019selective, huang2020interpretable} typically produce soft attention rather than hard decisions, which cannot reduce computation costs. Our approach can be more practical and effective benefiting from our careful designs.

\vspace{-0.5\baselineskip}
\subsection{Optimization}
The proposed network is optimized by the following loss function:
\begin{equation} \label{loss}
    \mathcal{L} = L_{ce}\left(p, l\right) + \lambda\sum_{k=1}^{K}d^{k},
\end{equation}
where $L_{ce}$ denotes the cross-entropy loss. $p$ and $l$ are the prediction logits and ground truth label. $d^{k}$ denotes the binarized discrete decision as in Eq. \ref{d_c_generation}. The second item in Eq. \ref{loss} encourages fewer regions to be selected, thus avoiding the redundant information and reducing computation costs. $\lambda$ is a hyper-parameter to control the trade-off between the two loss items.

\begin{table}[t]
    \centering
    \begin{adjustbox}{width=1.0\columnwidth,center}
    \renewcommand{\arraystretch}{0.9}
    \setlength\tabcolsep{18pt}
    \begin{tabular}{l | c | c}
    \toprule
    Method & Backbone & Accuracy\\
    \midrule
    ResNet50 \cite{he2016deep} & ResNet50 & 84.5 \\
    ResNet101 \cite{he2016deep} & ResNet101 & 85.5\\
    DenseNet161 \cite{huang2017densely} & DenseNet161 & 85.5\\
    \midrule
    PC-Dense161 \cite{dubey2018pairwise} & DenseNet161 & 86.9\\
    NTSNet \cite{yang2018learning} & ResNet50 & 87.5\\
    Cross-X \cite{luo2019cross} & ResNet50 & 87.7\\
    DCL \cite{chen2019destruction} & ResNet50 & 87.8\\
    S3N \cite{ding2019selective} & ResNet50 & 88.5 \\
    iSQRT-COV \cite{li2018towards} & ResNet101 & 88.7\\
    GaRD \cite{zhao2021graph} & ResNet50 & 89.6\\
    APINet \cite{zhuang2020learning} & DenseNet161 & 90.0\\
    \midrule
     \rowcolor{mygray} \textbf{Ours} & ResNet50 & 90.8\\
     \rowcolor{mygray} \textbf{Ours} & ResNet101 & 91.3\\
     \rowcolor{mygray} \textbf{Ours} & DenseNet161 & 91.5\\
     \bottomrule
    \end{tabular}
    \end{adjustbox}
    \caption{Comparison results on CUB-200-2011 dataset. }
    \label{cub}
\vspace{-1\baselineskip}
\end{table}

\begin{table*}[t]
	\hspace{0.3cm}
	\begin{minipage}[t]{0.63\columnwidth}
        \begin{adjustbox}{width=1.0\columnwidth,center}
        \begin{tabular}{l | c | c}
        \toprule
        Method & Backbone & Accuracy\\
        \midrule
        ResNet50 \cite{he2016deep} & ResNet50 & 82.2 \\
        ResNet101 \cite{he2016deep} & ResNet101 & 82.9\\
        DenseNet161 \cite{huang2017densely} & DenseNet161 & 83.1\\
        \midrule
        Cross-X \cite{luo2019cross} & ResNet50 & 86.4\\
        GaRD \cite{zhao2021graph} & ResNet50 & 88.0\\
        APINet \cite{zhuang2020learning} & DensetNet101 & 88.1\\
        \midrule
        \rowcolor{mygray} \textbf{Ours} & ResNet50 & 89.5\\
        \rowcolor{mygray} \textbf{Ours} & ResNet101 & 90.0\\
        \rowcolor{mygray} \textbf{Ours} & DenseNet161 & 90.4\\
         \bottomrule
        \end{tabular}
        \end{adjustbox}
		\caption{Comparison results on NA-bird dataset.}
		\label{nabird}
	\end{minipage}
	\hspace{0.3cm}
	\begin{minipage}[t]{0.63\columnwidth}
        \begin{adjustbox}{width=1.0\columnwidth,center}
		\centering
            \begin{tabular}{l | c | c}
                \toprule
                Method & Backbone & Accuracy\\
                \midrule
                ResNet50 \cite{he2016deep} & ResNet50 & 80.1 \\
                ResNet101 \cite{he2016deep} & ResNet101 & 80.6\\
                DenseNet161 \cite{huang2017densely} & DenseNet161 & 80.9\\
                \midrule
                PC-Dense161 \cite{dubey2018pairwise} & DenseNet161 & 82.8\\
                Cross-X \cite{luo2019cross} & ResNet50 & 88.9\\
                APINet \cite{zhuang2020learning} & ResNet101 & 90.3\\
                \midrule
                 \rowcolor{mygray} \textbf{Ours} & ResNet50 & 91.0\\
                 \rowcolor{mygray} \textbf{Ours} & ResNet101 & 91.7\\
                 \rowcolor{mygray} \textbf{Ours} & DenseNet161 & 92.1\\
                 \bottomrule
                \end{tabular}
                \end{adjustbox}
		\caption{Comparison results on Stanford Dogs dataset.}
		\label{dogs}
	\end{minipage}
    \hspace{0.3cm}
	\begin{minipage}[t]{0.63\columnwidth}
		\begin{adjustbox}{width=1.0\columnwidth,center}\centering
            \begin{tabular}{l | c | c}
                \toprule
                Method & Backbone & Accuracy\\
                \midrule
                ResNet50 \cite{he2016deep} & ResNet50 & 69.4 \\
                ResNet101 \cite{he2016deep} & ResNet101 & 72.0\\
                DenseNet161 \cite{huang2017densely} & DenseNet161 & 72.9\\
                \midrule
                B-CNN \cite{lin2015bilinear}& ResNet50 & 75.8\\
                Deep-TEN \cite{zhang2017deep} & ResNet50 & 76.1\\
                DEP \cite{xue2018deep} & ResNet50 & 82.7\\
                \midrule
                \rowcolor{mygray} \textbf{Ours} & ResNet50 & 86.4\\
                \rowcolor{mygray}\textbf{Ours} & ResNet101 & 87.0\\
                  \rowcolor{mygray} \textbf{Ours} & DenseNet161 & 87.5\\
                 \bottomrule
                \end{tabular}
                \end{adjustbox}
		\caption{Comparison results on GTOS-mobile dataset.}
		\label{gtos}
	\end{minipage}
\end{table*}

\vspace{-0.5\baselineskip}
\section{Experiments}
\subsection{Implementation Details}
Following \cite{zhuang2020learning}, we adopt different networks as the semantic branch, including ResNet50, ResNet101 and DenseNet161. We use SGD as the optimizer with the momentum of 0.9. The initial learning rate is set to 1e-4, which decays by 0.1 for every 20 epochs (overall 100 epochs). Batch size is 16 for all datasets. Following previous works \cite{zhao2021graph, lin2015bilinear}, we use random cropping of 448 × 448 in training and center crop at inference. We apply the commonly-used data augmentation strategies for training, including random scaling, left-right flipping and random cropping. $S$ indicating the size of regions that are zoomed in is set to 112. $N$ indicating the number of learnable Gabor filters is set to 128. $M$ indicating the number of quantization levels in QCO is set to 8. $\lambda$ in Eq. \ref{loss} is set to 0.2. Experiments are conducted on 2 NVIDIA TITAN V100 GPUs.

\subsection{Comparison with State-of-the-art}
We evaluate our method on four datasets, including three fine-grained datasets CUB-200-2011, NA-bird, Stanford Dogs and a terrain recognition dataset  GTOS-mobile. The comparison results on these datasets are presented in Tables \ref{cub}, \ref{nabird}, \ref{dogs}, and \ref{gtos}, respectively. Our method achieves the best performance when using DenseNet161 as the semantic branch, reaching accuracy scores of 91.5, 90.4, 92.1 and 87.5, outperforming the second-best method by 1.5, 2.3, 1.8 and 4.8, respectively. It is worth noting that even when using ResNet50, the least effective backbone among the three, our method remains superior to all other compared methods. This highlights the high effectiveness of our method. The advantage of our method is especially significant on the challenging terrain dataset GTOS-mobile. This is due to our method's ability to extract texture features, which are essential for the classification of visually-similar and closely-related terrain classes.

\begin{table}[t]
    \centering
    \begin{adjustbox}{width=1.0\columnwidth,center}
    \setlength\tabcolsep{14pt}
    \begin{tabular}{l | c c c}
    \toprule
    Method & Accuracy & Flops(G) & Params(M)\\
    \midrule
    Baseline (ResNet50) & 84.5 & 16.26 & 24.69\\
    \midrule
    + TB \& RSG & 90.8 & 20.72 & 32.41\\
    + TB (image-based manner) & 89.0 & 36.88 & 29.82\\
    + TB (patch-based manner) & 89.9 & 37.04 & 29.82\\
     \bottomrule
    \end{tabular}
    \end{adjustbox}
    \caption{Ablation results of different components in our method. TB and RSG refer to the texture branch and region selection gate, respectively.}
    \label{ablation_texture}
\vspace{-1\baselineskip}
\end{table}

\subsection{Ablation Study}
We perform ablation study to verify the effectiveness of our designs. Experiments in this part are conducted on CUB-200-2011 dataset with ResNet50 as the semantic branch. Due to paper length limitation, more ablation study results are presented in \textbf{supplementary materials}.

\noindent \textbf{Ablation of Different Components.} As shown in Fig. \ref{main_fig}, in addition to the basic CNN-based semantic branch, our method consists of two key components: a Gabor-filter-based texture branch (TB) and a region selection gate (RSG). We conduct experiments to validate the effect of these components and present the results in Table. \ref{ablation_texture}, which includes the validation accuracy, computation costs and parameter numbers of different settings. We set the network that only has a ResNet50 semantic branch to be the baseline. Our method with adding both texture branch \& region selection gate increases accuracy from 84.5 to 90.8. When the region selection gate is removed, extracting textures by feeding the whole image into texture branch (image-based manner) reduces accuracy to 89.0 and increases computation by 78\%; equally dividing image into $8\times 8$ patches and getting a feature from each patch (patch-based manner) reduces accuracy to 89.0 and increases computation by 79\%. The results demonstrate that the region selection gate can reduce computation and improve performance.

\begin{table}[t]
    \centering
    \begin{adjustbox}{width=1.0\columnwidth,center}
    \begin{tabular}{l | c c c}
    \toprule
    Method & Accuracy & Flops(G) & Params(M)\\
    \midrule
    Ours & 90.8 & 20.72 & 32.41\\
    \midrule
    Ours replacing Gabor filters with conv & 85.9 & 20.03 & 32.95\\
    Ours w/o parameter learning & 86.5 & 20.72 & 32.41\\
    Ours w/o parameter value constraints & 75.4 & 20.40 & 32.41\\
    Ours w/o high frequency enhancement & 88.1 & 20.40 & 32.41\\
     \bottomrule
    \end{tabular}
    \end{adjustbox}
    \caption{Ablation results of learnable Gabor filters.}
    \label{ablation_learn}
\vspace{-0.5\baselineskip}
\end{table}

\begin{figure}[t]
    \subfigure[Loss Curves]{
    \begin{minipage}[t]{0.45\linewidth}
    \centering
    \includegraphics[width=1\linewidth]{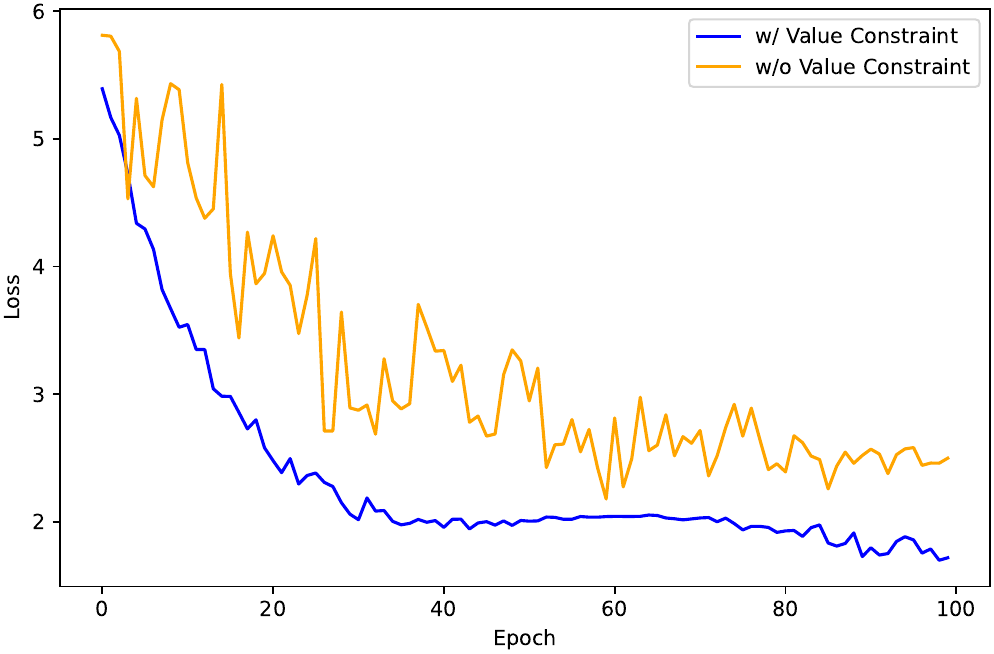}
    \end{minipage}%
    }%
    \subfigure[Accuracy Curves]{
    \begin{minipage}[t]{0.45\linewidth}
    \centering
    \includegraphics[width=1\linewidth]{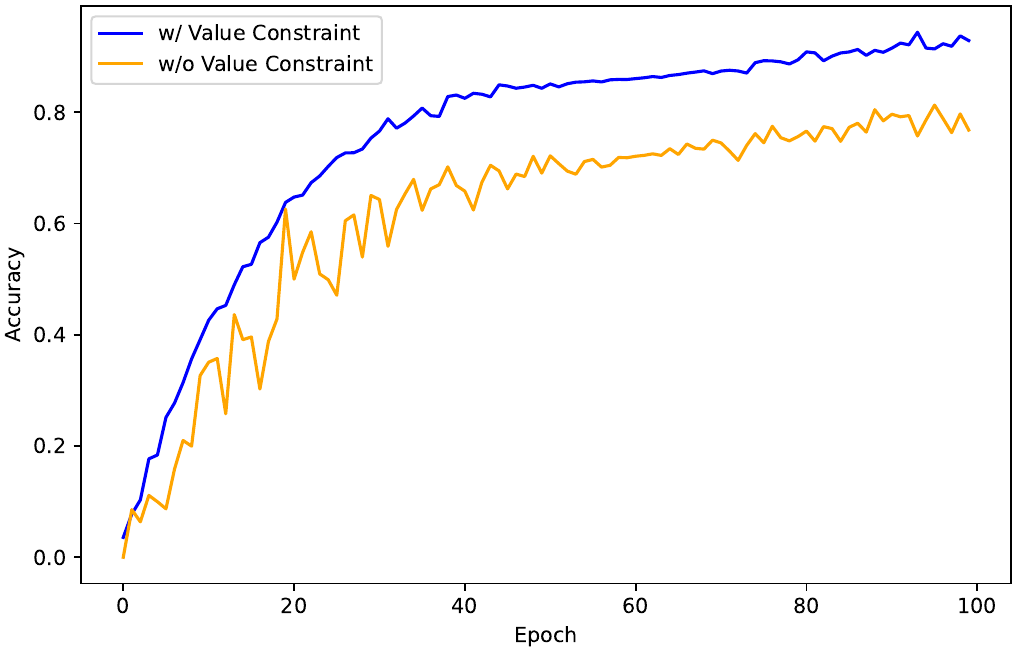}
    \end{minipage}%
    }%
    \caption{Loss curves (a) and accuracy curves (b) in different settings. {\color{blue} Blue} and {\color{orange} orange} lines indicate curves for training Gabor filters {\color{blue} w/} and {\color{orange} w/o} using parameter value constraints, respectively.}
    \label{curve}
\vspace{-0.7\baselineskip}
\end{figure}

\noindent \textbf{Ablation of Learnable Gabor Filters. }Results in this part are presented in Table. \ref{ablation_learn}. (1) First, we validate the effectiveness of Gabor filters. Replacing all Gabor filters by the $3\times 3$ convolution layers decreases accuracy by 4.9. This indicates that the improvement does not simply come from an increase in the number of parameters, but from the effectiveness of using Gabor filters for texture extraction. (2) Second, we evaluate the effectiveness of learning Gabor filter parameters in a supervised manner. We replace the learnable parameters with the manually set ones, where the setting rules are the same as \cite{tao2007general}. This reduces accuracy by 4.3, showing the effectiveness of supervised learning. (3) Third, we verify the necessity of using parameter value constraints. In Fig. \ref{curve}, we present the loss curves and validation accuracy curves for methods w/ and w/o using value constraints. Using value constraints results in a smoother loss decline curve and accuracy improvement curve. In contrast, free training causes violent loss and accuracy fluctuations, especially at the beginning stage of training, indicating that the training process is unstable. Compared to free training, using value constraints significantly improves validation accuracy by 15.4. (4) Last, we evaluate the high frequency enhancement strategy by setting two constraints for frequency parameter $W$. Without the strategy and directly constraining $W$ into its valid range reduces accuracy by 2.7. Results demonstrate the effectiveness of our designs for the learnable Gabor filter parameters. Also note, using the proposed learning methods requires no extra parameters and only increases computation very slightly.  

\noindent \textbf{Ablation of Statistical Feature Extractor.} We propose a statistical feature extractor to leverage texture information from the output of Gabor filters. The extractor consists of two parts: a Learnable Histogram Operator (LHO) and a Filter Correlation Module (FCM). We conduct experiments to validate the effect of these components and present the results in Table. \ref{ablation_extractor}. Removing LHO and FCM reduces accuracy from 90.8 to 87.0 and 89.2 respectively. Also note that, both LHO and FCM are lightweight, only requiring extra parameters of 4.26M and 3.56M, increasing flops by 15\% and 6\% respectively. Results show the high effectiveness and efficiency of LHO and FCM.

\begin{table}[t]
    \centering
    \begin{adjustbox}{width=1.0\columnwidth,center}
    \setlength\tabcolsep{18pt}
    \begin{tabular}{l | c c c}
    \toprule
    Method & Accuracy & Flops(G) & Params(M)\\
    \midrule
    Ours & 90.8 & 20.72 & 32.41\\
    \midrule
    Ours w/o LHO  & 87.0 & 18.03 & 28.25\\
    Ours w/o FCM & 89.2 & 19.55 & 28.85\\
     \bottomrule
    \end{tabular}
    \end{adjustbox}
    \caption{Ablation results of statistical feature extractor. }
    \label{ablation_extractor}
\vspace{-1\baselineskip}
\end{table}

\noindent \textbf{Ablation of LHO. } In Table. \ref{ablation_qco}, we further evaluate the designs in LHO. First, we replace LHO with a hand-crafted statistical feature composed of mean, variance, maximum and minimum values, which reduces accuracy by 2.9. This demonstrates that LHO can produce statistical features that are more effective than hand-crafted ones widely used by previous methods. Next, we remove position information $\mathbf{P}^{m}$ in Eq. \ref{encode_position} and multi-head attention in Eq. \ref{correlate_level}, which reduces accuracy by 1.5 and 2.1 respectively, showing the importance of encoding position features and correlating different levels in LHO. Finally, we move LHO from texture branch to CNN-based semantic branch. This reduces accuracy by 4.6, showing that the effectiveness of LHO relies on Gabor filters to extract texture-related signals.

\subsection{Visualization}
In Fig. \ref{visual}, we provide a visualization of the outputs obtained from learned Gabor filters. More specifically, Fig. \ref{visual} (c) and Fig. \ref{visual} (d) show the average output of all high-frequency and low-frequency Gabor filters, respectively. As can be observed, the high-frequency filters primarily capture information of undulating areas such as speckles and ripples, whereas the low-frequency filters primarily capture information related to smooth changing areas. Both these two kinds of information are critical for recognition. By exploiting sufficient and balanced multi-frequency features through the carefully-designed learnable Gabor filters, our method can leverage comprehensive information for effective fine-grained recognition. Due to space limitation, more visualization results are presented in \textbf{supplementary}. 

\begin{figure}[t]
    \subfigure[\tiny Image]{
    \begin{minipage}[t]{0.3\linewidth}
    \centering
    \includegraphics[width=0.82\linewidth]{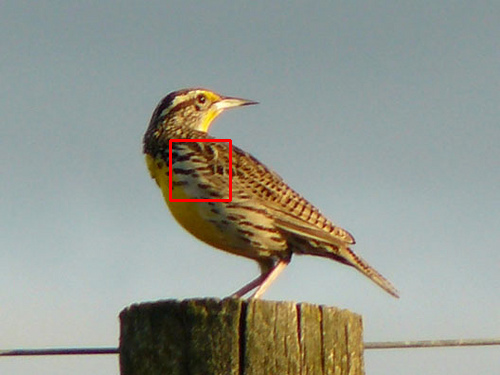}\\
    \includegraphics[width=0.82\linewidth]{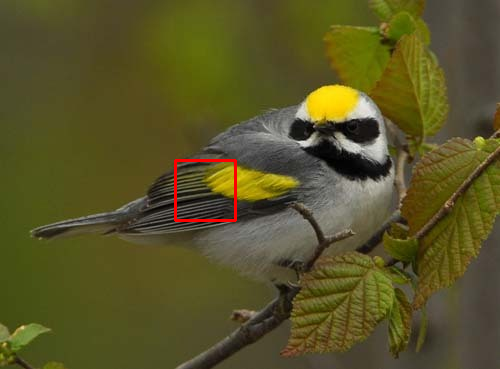}
    \end{minipage}%
    }%
    \subfigure[{\tiny Region}]{
    \begin{minipage}[t]{0.2\linewidth}
    \centering
    \includegraphics[width=0.9\linewidth]{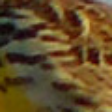}\\
    \includegraphics[width=0.9\linewidth]{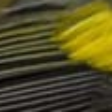}
    \end{minipage}%
    }%
    \subfigure[\tiny High-$W$ Filters]{
    \begin{minipage}[t]{0.2\linewidth}
    \centering
    \includegraphics[width=0.9\linewidth]{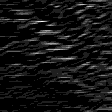}\\
    \includegraphics[width=0.9\linewidth]{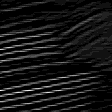}
    \end{minipage}%
    }%
    \subfigure[\tiny Low-$W$ Filters]{
    \begin{minipage}[t]{0.2\linewidth}
    \centering
    \includegraphics[width=0.9\linewidth]{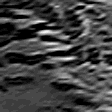}\\
    \includegraphics[width=0.9\linewidth]{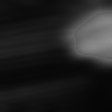}
    \end{minipage}%
    }%
    \caption{Visualization of Gabor filters. (a), (b), (c) and (d) present the original images, the selected regions, average output of all high-frequency and low-frequency Gabor filters, respectively.}
    \label{visual}
\vspace{-0.7\baselineskip}
\end{figure}

\begin{table}[t]
    \centering
    \begin{adjustbox}{width=1.0\columnwidth,center}
    \setlength\tabcolsep{12pt}
    \begin{tabular}{l | c c c}
    \toprule
    Method & Accuracy & Flops(G) & Params(M)\\
    \midrule
    Ours & 90.8 & 20.72 & 32.41\\
    \midrule
    Ours w/o LHO w/ HSF & 87.9 & 18.16 & 28.25\\
    Ours w/o $\mathbf{P}^{m}$ in Eq. \ref{encode_position} & 89.3 & 18.50 & 30.93\\
    Ours w/o MHA in Eq.\ref{correlate_level} & 88.7 & 19.49 & 29.73\\
    Ours w/o LHO in TB w/ LHO in CNN & 86.2 & 22.85 & 32.90\\
     \bottomrule
    \end{tabular}
    \end{adjustbox}
    \caption{Ablation results of LHO. HSF refers to a hand-crafted statistical feature composed of mean, variance, maximum and minimum values. MHA denotes multi-head attention. TB denotes texture branch.}
    \label{ablation_qco}
\vspace{-1\baselineskip}
\end{table}

\subsection{Discussion of Computation and Parameter}
Compared to the ResNet50 baseline, our method only increases computation and parameter usage slightly. Specifically, on CUB-200-2011 dataset, ResNet50 consumes 16.26G flops of computation, requiring 24.69M parameters, reaching accuracy of 84.5. Our method consumes 20.72G flops of computation, requiring 32.41M parameters, reaching accuracy of 90.8. Compared to ResNet50, our method can improve accuracy significantly (+6.3) with only increasing computation and parameter usage by 27\% and 31\% respectively. This demonstrates the high effectiveness and high efficiency of our method with outstanding performance and reasonable computation costs. 

\vspace{-0.5\baselineskip}
\section{Conclusion}
\vspace{-0.5\baselineskip}
This paper presents a novel network that extracts texture features to facilitate fine-grained recognition using Gabor filters. We make the Gabor filter parameters to be automatically learned in a stable and effective manner through our carefully-designed value constraint strategies. We also propose a statistical feature extractor and a region selection gate, allowing informative and effective Gabor texture features to be extracted with very few extra computation costs. Our method is effective, as shown by the extensive experiments with state-of-the-art (SOTA) performance.\\

\noindent \textbf{Acknowledgement}
This research is supported by the National Research Foundation, Singapore under its AI Singapore Programme (AISG Award No: AISG2-PhD-2021-08-006), Singapore Ministry of Education (MOE) AcRF Tier 2 under Grant MOE-T2EP20222-0009, the National Research Foundation Singapore through AI Singapore Programme under Grant AISG-100E-2020-065, and SUTD SKI Project under Grant SKI 2021\_02\_06. 

{\small
\bibliographystyle{ieee_fullname}
\bibliography{egbib}

\begin{thebibliography}{10}\itemsep=-1pt

\bibitem{arrospide2013log}
Jon Arrospide and Luis Salgado.
\newblock Log-gabor filters for image-based vehicle verification.
\newblock {\em IEEE Transactions on Image Processing}, 22(6):2286--2295, 2013.

\bibitem{branson2014bird}
Steve Branson, Grant Van~Horn, Serge Belongie, and Pietro Perona.
\newblock Bird species categorization using pose normalized deep convolutional
  nets.
\newblock {\em arXiv preprint arXiv:1406.2952}, 2014.

\bibitem{chen2019destruction}
Yue Chen, Yalong Bai, Wei Zhang, and Tao Mei.
\newblock Destruction and construction learning for fine-grained image
  recognition.
\newblock In {\em Proceedings of the IEEE/CVF conference on computer vision and
  pattern recognition}, pages 5157--5166, 2019.

\bibitem{clausi2005design}
David~A Clausi and Huang Deng.
\newblock Design-based texture feature fusion using gabor filters and
  co-occurrence probabilities.
\newblock {\em IEEE transactions on image processing}, 14(7):925--936, 2005.

\bibitem{daugman1980two}
John~G Daugman.
\newblock Two-dimensional spectral analysis of cortical receptive field
  profiles.
\newblock {\em Vision research}, 20(10):847--856, 1980.

\bibitem{daugman1985uncertainty}
John~G Daugman.
\newblock Uncertainty relation for resolution in space, spatial frequency, and
  orientation optimized by two-dimensional visual cortical filters.
\newblock {\em JOSA A}, 2(7):1160--1169, 1985.

\bibitem{ding2019selective}
Yao Ding, Yanzhao Zhou, Yi Zhu, Qixiang Ye, and Jianbin Jiao.
\newblock Selective sparse sampling for fine-grained image recognition.
\newblock In {\em Proceedings of the IEEE/CVF International Conference on
  Computer Vision}, pages 6599--6608, 2019.

\bibitem{dosovitskiy2020image}
Alexey Dosovitskiy, Lucas Beyer, Alexander Kolesnikov, Dirk Weissenborn,
  Xiaohua Zhai, Thomas Unterthiner, Mostafa Dehghani, Matthias Minderer, Georg
  Heigold, Sylvain Gelly, et~al.
\newblock An image is worth 16x16 words: Transformers for image recognition at
  scale.
\newblock {\em arXiv preprint arXiv:2010.11929}, 2020.

\bibitem{dubey2018pairwise}
Abhimanyu Dubey, Otkrist Gupta, Pei Guo, Ramesh Raskar, Ryan Farrell, and
  Nikhil Naik.
\newblock Pairwise confusion for fine-grained visual classification.
\newblock In {\em Proceedings of the European conference on computer vision
  (ECCV)}, pages 70--86, 2018.

\bibitem{dubey2018maximum}
Abhimanyu Dubey, Otkrist Gupta, Ramesh Raskar, and Nikhil Naik.
\newblock Maximum-entropy fine grained classification.
\newblock {\em Advances in neural information processing systems}, 31, 2018.

\bibitem{fu2017look}
Jianlong Fu, Heliang Zheng, and Tao Mei.
\newblock Look closer to see better: Recurrent attention convolutional neural
  network for fine-grained image recognition.
\newblock In {\em Proceedings of the IEEE conference on computer vision and
  pattern recognition}, pages 4438--4446, 2017.

\bibitem{fu2022panoptic}
Xiao Fu, Shangzhan Zhang, Tianrun Chen, Yichong Lu, Lanyun Zhu, Xiaowei Zhou,
  Andreas Geiger, and Yiyi Liao.
\newblock Panoptic nerf: 3d-to-2d label transfer for panoptic urban scene
  segmentation.
\newblock In {\em 2022 International Conference on 3D Vision (3DV)}, pages
  1--11. IEEE, 2022.

\bibitem{gao2016compact}
Yang Gao, Oscar Beijbom, Ning Zhang, and Trevor Darrell.
\newblock Compact bilinear pooling.
\newblock In {\em Proceedings of the IEEE conference on computer vision and
  pattern recognition}, pages 317--326, 2016.

\bibitem{ge2019weakly}
Weifeng Ge, Xiangru Lin, and Yizhou Yu.
\newblock Weakly supervised complementary parts models for fine-grained image
  classification from the bottom up.
\newblock In {\em Proceedings of the IEEE/CVF Conference on Computer Vision and
  Pattern Recognition}, pages 3034--3043, 2019.

\bibitem{girshick2014rich}
Ross Girshick, Jeff Donahue, Trevor Darrell, and Jitendra Malik.
\newblock Rich feature hierarchies for accurate object detection and semantic
  segmentation.
\newblock In {\em Proceedings of the IEEE conference on computer vision and
  pattern recognition}, pages 580--587, 2014.

\bibitem{gonzales1987digital}
Rafael~C Gonzales and Paul Wintz.
\newblock {\em Digital image processing}.
\newblock Addison-Wesley Longman Publishing Co., Inc., 1987.

\bibitem{he2016deep}
Kaiming He, Xiangyu Zhang, Shaoqing Ren, and Jian Sun.
\newblock Deep residual learning for image recognition.
\newblock In {\em Proceedings of the IEEE conference on computer vision and
  pattern recognition}, pages 770--778, 2016.

\bibitem{hu2020gabor}
Xiao-dong Hu, Xin-qing Wang, Fan-jie Meng, Xia Hua, Yu-ji Yan, Yu-yang Li, Jing
  Huang, and Xun-lin Jiang.
\newblock Gabor-cnn for object detection based on small samples.
\newblock {\em Defence Technology}, 16(6):1116--1129, 2020.

\bibitem{huang2017densely}
Gao Huang, Zhuang Liu, Laurens Van Der~Maaten, and Kilian~Q Weinberger.
\newblock Densely connected convolutional networks.
\newblock In {\em Proceedings of the IEEE conference on computer vision and
  pattern recognition}, pages 4700--4708, 2017.

\bibitem{huang2016part}
Shaoli Huang, Zhe Xu, Dacheng Tao, and Ya Zhang.
\newblock Part-stacked cnn for fine-grained visual categorization.
\newblock In {\em Proceedings of the IEEE conference on computer vision and
  pattern recognition}, pages 1173--1182, 2016.

\bibitem{huang2020interpretable}
Zixuan Huang and Yin Li.
\newblock Interpretable and accurate fine-grained recognition via region
  grouping.
\newblock In {\em Proceedings of the IEEE/CVF Conference on Computer Vision and
  Pattern Recognition}, pages 8662--8672, 2020.

\bibitem{jain1997object}
Anil~K Jain, Nalini~K Ratha, and Sridhar Lakshmanan.
\newblock Object detection using gabor filters.
\newblock {\em Pattern recognition}, 30(2):295--309, 1997.

\bibitem{ji2020attention}
Ruyi Ji, Longyin Wen, Libo Zhang, Dawei Du, Yanjun Wu, Chen Zhao, Xianglong
  Liu, and Feiyue Huang.
\newblock Attention convolutional binary neural tree for fine-grained visual
  categorization.
\newblock In {\em Proceedings of the IEEE/CVF Conference on Computer Vision and
  Pattern Recognition}, pages 10468--10477, 2020.

\bibitem{kaiser2018discrete}
{\L}ukasz Kaiser and Samy Bengio.
\newblock Discrete autoencoders for sequence models.
\newblock {\em arXiv preprint arXiv:1801.09797}, 2018.

\bibitem{kaiser2018fast}
Lukasz Kaiser, Samy Bengio, Aurko Roy, Ashish Vaswani, Niki Parmar, Jakob
  Uszkoreit, and Noam Shazeer.
\newblock Fast decoding in sequence models using discrete latent variables.
\newblock In {\em International Conference on Machine Learning}, pages
  2390--2399. PMLR, 2018.

\bibitem{kong2017low}
Shu Kong and Charless Fowlkes.
\newblock Low-rank bilinear pooling for fine-grained classification.
\newblock In {\em Proceedings of the IEEE conference on computer vision and
  pattern recognition}, pages 365--374, 2017.

\bibitem{li2018towards}
Peihua Li, Jiangtao Xie, Qilong Wang, and Zilin Gao.
\newblock Towards faster training of global covariance pooling networks by
  iterative matrix square root normalization.
\newblock In {\em Proceedings of the IEEE conference on computer vision and
  pattern recognition}, pages 947--955, 2018.

\bibitem{lin2015bilinear}
Tsung-Yu Lin, Aruni RoyChowdhury, and Subhransu Maji.
\newblock Bilinear cnn models for fine-grained visual recognition.
\newblock In {\em Proceedings of the IEEE international conference on computer
  vision}, pages 1449--1457, 2015.

\bibitem{liu2004gabor}
Chengjun Liu.
\newblock Gabor-based kernel pca with fractional power polynomial models for
  face recognition.
\newblock {\em IEEE transactions on pattern analysis and machine intelligence},
  26(5):572--581, 2004.

\bibitem{liu2003independent}
Chengjun Liu and Harry Wechsler.
\newblock Independent component analysis of gabor features for face
  recognition.
\newblock {\em IEEE transactions on Neural Networks}, 14(4):919--928, 2003.

\bibitem{liu2021swin}
Ze Liu, Yutong Lin, Yue Cao, Han Hu, Yixuan Wei, Zheng Zhang, Stephen Lin, and
  Baining Guo.
\newblock Swin transformer: Hierarchical vision transformer using shifted
  windows.
\newblock In {\em Proceedings of the IEEE/CVF international conference on
  computer vision}, pages 10012--10022, 2021.

\bibitem{luan2018gabor}
Shangzhen Luan, Chen Chen, Baochang Zhang, Jungong Han, and Jianzhuang Liu.
\newblock Gabor convolutional networks.
\newblock {\em IEEE Transactions on Image Processing}, 27(9):4357--4366, 2018.

\bibitem{luo2019theory}
Tao Luo, Zheng Ma, Zhi-Qin~John Xu, and Yaoyu Zhang.
\newblock Theory of the frequency principle for general deep neural networks.
\newblock {\em arXiv preprint arXiv:1906.09235}, 2019.

\bibitem{luo2019cross}
Wei Luo, Xitong Yang, Xianjie Mo, Yuheng Lu, Larry~S Davis, Jun Li, Jian Yang,
  and Ser-Nam Lim.
\newblock Cross-x learning for fine-grained visual categorization.
\newblock In {\em Proceedings of the IEEE/CVF international conference on
  computer vision}, pages 8242--8251, 2019.

\bibitem{oh2018analytic}
Beom-Seok Oh, Kar-Ann Toh, Andrew Beng~Jin Teoh, and Zhiping Lin.
\newblock An analytic gabor feedforward network for single-sample and
  pose-invariant face recognition.
\newblock {\em IEEE Transactions on Image Processing}, 27(6):2791--2805, 2018.

\bibitem{simon2015neural}
Marcel Simon and Erik Rodner.
\newblock Neural activation constellations: Unsupervised part model discovery
  with convolutional networks.
\newblock In {\em Proceedings of the IEEE international conference on computer
  vision}, pages 1143--1151, 2015.

\bibitem{steiglitz2020digital}
Kenneth Steiglitz.
\newblock {\em Digital Signal Processing Primer}.
\newblock Courier Dover Publications, 2020.

\bibitem{sun2018multi}
Ming Sun, Yuchen Yuan, Feng Zhou, and Errui Ding.
\newblock Multi-attention multi-class constraint for fine-grained image
  recognition.
\newblock In {\em Proceedings of the european conference on computer vision
  (ECCV)}, pages 805--821, 2018.

\bibitem{sun2005road}
Zehang Sun, George Bebis, and Ronald Miller.
\newblock On-road vehicle detection using evolutionary gabor filter
  optimization.
\newblock {\em IEEE Transactions on Intelligent Transportation Systems},
  6(2):125--137, 2005.

\bibitem{tao2007general}
Dacheng Tao, Xuelong Li, Xindong Wu, and Stephen~J Maybank.
\newblock General tensor discriminant analysis and gabor features for gait
  recognition.
\newblock {\em IEEE transactions on pattern analysis and machine intelligence},
  29(10):1700--1715, 2007.

\bibitem{wang2020weakly}
Zhihui Wang, Shijie Wang, Shuhui Yang, Haojie Li, Jianjun Li, and Zezhou Li.
\newblock Weakly supervised fine-grained image classification via guassian
  mixture model oriented discriminative learning.
\newblock In {\em Proceedings of the IEEE/CVF Conference on Computer Vision and
  Pattern Recognition}, pages 9749--9758, 2020.

\bibitem{weber2001quadratic}
David~M Weber and David~P Casasent.
\newblock Quadratic gabor filters for object detection.
\newblock {\em IEEE Transactions on Image Processing}, 10(2):218--230, 2001.

\bibitem{wei2018mask}
Xiu-Shen Wei, Chen-Wei Xie, Jianxin Wu, and Chunhua Shen.
\newblock Mask-cnn: Localizing parts and selecting descriptors for fine-grained
  bird species categorization.
\newblock {\em Pattern Recognition}, 76:704--714, 2018.

\bibitem{xiao2015application}
Tianjun Xiao, Yichong Xu, Kuiyuan Yang, Jiaxing Zhang, Yuxin Peng, and Zheng
  Zhang.
\newblock The application of two-level attention models in deep convolutional
  neural network for fine-grained image classification.
\newblock In {\em Proceedings of the IEEE conference on computer vision and
  pattern recognition}, pages 842--850, 2015.

\bibitem{xu2019frequency}
Zhi-Qin~John Xu, Yaoyu Zhang, Tao Luo, Yanyang Xiao, and Zheng Ma.
\newblock Frequency principle: Fourier analysis sheds light on deep neural
  networks.
\newblock {\em arXiv preprint arXiv:1901.06523}, 2019.

\bibitem{xu2019training2}
Zhi-Qin~John Xu, Yaoyu Zhang, and Yanyang Xiao.
\newblock Training behavior of deep neural network in frequency domain.
\newblock In {\em Neural Information Processing: 26th International Conference,
  ICONIP 2019, Sydney, NSW, Australia, December 12--15, 2019, Proceedings, Part
  I 26}, pages 264--274. Springer, 2019.

\bibitem{xue2018deep}
Jia Xue, Hang Zhang, and Kristin Dana.
\newblock Deep texture manifold for ground terrain recognition.
\newblock In {\em Proceedings of the IEEE Conference on Computer Vision and
  Pattern Recognition}, pages 558--567, 2018.

\bibitem{yang2018learning}
Ze Yang, Tiange Luo, Dong Wang, Zhiqiang Hu, Jun Gao, and Liwei Wang.
\newblock Learning to navigate for fine-grained classification.
\newblock In {\em Proceedings of the European conference on computer vision
  (ECCV)}, pages 420--435, 2018.

\bibitem{yu2018hierarchical}
Chaojian Yu, Xinyi Zhao, Qi Zheng, Peng Zhang, and Xinge You.
\newblock Hierarchical bilinear pooling for fine-grained visual recognition.
\newblock In {\em Proceedings of the European conference on computer vision
  (ECCV)}, pages 574--589, 2018.

\bibitem{yuan2022adaptive}
Ye Yuan, Li-Na Wang, Guoqiang Zhong, Wei Gao, Wencong Jiao, Junyu Dong, Biao
  Shen, Dongdong Xia, and Wei Xiang.
\newblock Adaptive gabor convolutional networks.
\newblock {\em Pattern Recognition}, 124:108495, 2022.

\bibitem{zhang2017deep}
Hang Zhang, Jia Xue, and Kristin Dana.
\newblock Deep ten: Texture encoding network.
\newblock In {\em Proceedings of the IEEE conference on computer vision and
  pattern recognition}, pages 708--717, 2017.

\bibitem{zhang2023painting}
Shangzhan Zhang, Sida Peng, Tianrun Chen, Linzhan Mou, Haotong Lin, Kaicheng
  Yu, Yiyi Liao, and Xiaowei Zhou.
\newblock Painting 3d nature in 2d: View synthesis of natural scenes from a
  single semantic mask.
\newblock In {\em Proceedings of the IEEE/CVF Conference on Computer Vision and
  Pattern Recognition}, pages 8518--8528, 2023.

\bibitem{zhao2021fractional}
Xudong Zhao, Ran Tao, Wei Li, Wilfried Philips, and Wenzhi Liao.
\newblock Fractional gabor convolutional network for multisource remote sensing
  data classification.
\newblock {\em IEEE Transactions on Geoscience and Remote Sensing}, 60:1--18,
  2021.

\bibitem{zhao2021graph}
Yifan Zhao, Ke Yan, Feiyue Huang, and Jia Li.
\newblock Graph-based high-order relation discovery for fine-grained
  recognition.
\newblock In {\em Proceedings of the IEEE/CVF Conference on Computer Vision and
  Pattern Recognition}, pages 15079--15088, 2021.

\bibitem{zheng2019learning}
Heliang Zheng, Jianlong Fu, Zheng-Jun Zha, and Jiebo Luo.
\newblock Learning deep bilinear transformation for fine-grained image
  representation.
\newblock {\em Advances in Neural Information Processing Systems}, 32, 2019.

\bibitem{zheng2019looking}
Heliang Zheng, Jianlong Fu, Zheng-Jun Zha, and Jiebo Luo.
\newblock Looking for the devil in the details: Learning trilinear attention
  sampling network for fine-grained image recognition.
\newblock In {\em Proceedings of the IEEE/CVF Conference on Computer Vision and
  Pattern Recognition}, pages 5012--5021, 2019.

\bibitem{zhu2023continual}
Lanyun Zhu, Tianrun Chen, Jianxiong Yin, Simon See, and Jun Liu.
\newblock Continual semantic segmentation with automatic memory sample
  selection.
\newblock In {\em Proceedings of the IEEE/CVF Conference on Computer Vision and
  Pattern Recognition}, pages 3082--3092, 2023.

\bibitem{zhu2021learning}
Lanyun Zhu, Deyi Ji, Shiping Zhu, Weihao Gan, Wei Wu, and Junjie Yan.
\newblock Learning statistical texture for semantic segmentation.
\newblock In {\em Proceedings of the IEEE/CVF Conference on Computer Vision and
  Pattern Recognition}, pages 12537--12546, 2021.

\bibitem{zhuang2020learning}
Peiqin Zhuang, Yali Wang, and Yu Qiao.
\newblock Learning attentive pairwise interaction for fine-grained
  classification.
\newblock In {\em Proceedings of the AAAI conference on artificial
  intelligence}, volume~34, pages 13130--13137, 2020.

\end{thebibliography}
}
\clearpage

\appendix 

The supplementary materials are arranged as follows. In Sec. \ref{detail}, additional details are presented to provide a more comprehensive understanding of our method. In Sec. \ref{experiment}, further experimental results are included to validate the effectiveness of our approach. In Sec. \ref{visual} and \ref{analysis}, additional visualization results and analysis are provided to offer deeper insights into the functioning of our method.

\section{More Details of Method} \label{detail}
\subsection{Derivation of Parameter Valid Ranges}
In Sec 3.3 of the text, we propose a constraint to ensure the stability of model training. This constraint limits the values of Gabor filter parameters into their valid ranges, which can be derived based on the properties of Gabor filters and digital images in different domains. According to the periodicity of angles, the orientation parameter has a valid range of $[0, \pi]$. As for the other parameters, we provide a detailed derivation of their valid ranges in the subsequent sections.\\

\noindent \textbf{Scale Parameter $[\sigma_{x}, \sigma_{y}]$.} $[\sigma_{x}, \sigma_{y}]$ are the scale parameters that determine the filter effective size in both spatial and frequency domains. In spatial domain, a Gaussian function modulates the sinusoidal plane wave, which is defined in the infinite signal space to satisfy its mathematical properties \cite{steiglitz2020digital}. However, an image is a finite length signal in the spatial domain, with its valid signal zone determined by the image width $S$. Specifically for an image with size $S \times S$, the valid zone for each axis can be expressed as $[-0.5S, 0.5S]$ with the image center as the coordinate origin. Previous research \cite{steiglitz2020digital} indicates that directly applying infinite-length-defined filters to the finite-length image zone would cause mathematical deficiency, which could limit the effectiveness of Gabor filters due to waveform distortion. To alleviate the problem, we propose a solution to concentrate most of the Gabor filter energy within the finite signal zone. This ensures that only a small amount of filter energy spills over the finite signal, minimizing the negative effects of using infinite-length-defined filters on finite-length images. Specifically for a Gaussian with mean $\mathbf{\mu}$ and variance $\sigma$, we constrain $[\mu-\alpha\sigma, \mu+\alpha\sigma]$ to fall in the valid signal zone of the image. $\alpha$ is a hyper-parameter to control the energy concentration degree. According to the experimental results shown in Table. \ref{ablation_alpha}, we choose $2.5$ to be the optimal value for $\alpha$. By doing so, 98.76\% of the Gaussian energy can be subtended, and only 1.24\% of filter energy spills out of the image signal, whose negative effect is negligible. Based on the above analysis, we derive the spatial-wise constraints for parameters $[\sigma_{x}, \sigma_{y}]$ as follows:
\begin{equation}
\label{spatial_constraint}
\left\{
\begin{aligned}
    & [-0.25\sigma_{x}, 0.25\sigma_{x}] \subseteq [-0.5S,0.5S]\\
    & [-0.25\sigma_{y}, 0.25\sigma_{y}] \subseteq [-0.5S,0.5S]
\end{aligned}
\right.
\end{equation}
We further analyze the frequency-wise constraints for $[\sigma_{x}, \sigma_{y}]$. We perform a Fourier transform on Eq. 1 of text and get the frequency-wise expression of Gabor filters as follows:
\begin{equation}
\label{freq_gabor}
    G\left(u, v\right) = \exp\left[-\frac{1}{2}\left(\left(4\pi^{2}\sigma_{x}^{2}\left(u-W\right)^{2}\right)+4\pi^{2}\sigma_{y}^{2}v^{2}\right)\right],
\end{equation}
\begin{table}[t]
    \centering
    \begin{adjustbox}{width=0.8\columnwidth,center}
    \setlength\tabcolsep{20pt}
    \begin{tabular}{l c | c}
    \toprule
    $\alpha$ & Subtended Energy & Accuracy\\
    \midrule
    1.0 & 68.27\% & 85.8\\
    1.5 & 86.64\% & 89.0\\
    2.0 & 95.45\% & 90.2\\
    2.5 & 98.76\% & 90.8\\
    3.0 & 99.73\% & 90.3\\
    3.5 & 99.95\% & 90.0\\
     \bottomrule
    \end{tabular}
    \end{adjustbox}
    \caption{Ablation results of hyper-parameter $\alpha$ for constraining Gabor filter parameters. When the value of $\alpha$ is too small, the percentage of subtended energy is also low. This leads to a large amount of filter energy spilling out of the effective signal zone, which in turn negatively impacts training stability. Conversely, if $\alpha$ is too large, the filter parameters may be constrained to a small range, leading to a loss of information across certain frequencies. Experimental results show that the optimal value for $\alpha$ is 2.5. This choice strikes a balance between training stability and the availability of sufficient multi-frequency information. }
    \label{ablation_alpha}
\end{table}
As can be observed from Eq. \ref{freq_gabor}, in frequency domain, the Gabor filter also contains a Gaussian with mean $\{W, 0\}$ and variance $\{\frac{1}{2\pi \sigma_{x}}, \frac{1}{2\pi \sigma_{y}}\}$ to control its effective size. The valid signal zone in frequency domain can be derived according to Nyquist sampling theorem, which indicates that for a given sample rate $f_{s}$, perfect reconstruction is guaranteed possible when the frequency $|W|<\left(f_{s}/2\right)$, otherwise signal aliasing would happen. In an image, the sample rate equals 1 pixel, so any frequency component larger than 0.5 is distorted thus being invalid. This means the valid signal zone in frequency domain is $[-0.5, 0.5]$. Following the constraints in spatial domain, we subtend 98.76\% of the frequency-wise Gaussian energy into the valid signal zone to avoid distortion, getting the constraints in frequency domains as follows:
\begin{equation}
\label{frequency_domain_constraint}
\left\{
\begin{aligned}
    & [W - \frac{2.5}{2\pi\sigma_{x}}, W+\frac{2.5}{2\pi\sigma_{x}}] \subseteq [-0.5, 0.5]\\
    & [-\frac{2.5}{2\pi\sigma_{y}}, \frac{2.5}{2\pi\sigma_{y}}] \subseteq [-0.5, 0.5]
\end{aligned}
\right.
\end{equation}
Solving Eq. \ref{spatial_constraint} and Eq. \ref{frequency_domain_constraint}, we get $[\frac{5}{2\pi (1-2W)}, \frac{S}{5}]$ and $[\frac{5}{2\pi}, \frac{S}{5}]$ to be the valid ranges for $\sigma_{x}$ and $\sigma_{y}$ respectively.\\

\noindent \textbf{Frequency Parameter $W$.} We further analyze the valid range for the frequency parameter $W$. Due to the symmetry of image frequencies, any $W$ less than 0 is mirrored with its opposite number $-W$, so frequency components less than 0 are not considered and the lower bound for $W$ is set to 0. The upper bound can be derived from two constraints. First, according to Nyquist sampling theorem, frequency should be lower than 1/2 to avoid aliasing. Second, the upper bound of $\sigma_{x}$ should be higher than its lower bound. Mathematically, these constraints are formulated as follows:
\begin{equation}
\label{frequency_constraint}
\left\{
\begin{aligned}
    &W < 0.5\\
    &\frac{S}{5} > \frac{5}{2\pi \left(1-2W\right)} 
\end{aligned}
\right.
\end{equation}
Solving Eq.\ref{frequency_constraint}, we get $\frac{2\pi S-25}{4\pi S}$ to be the upper bound and $[0, \frac{2\pi S-25}{4\pi S}]$ to be the valid range of $W$.

\begin{figure}
    \centering
    \includegraphics[width=1.0\linewidth]{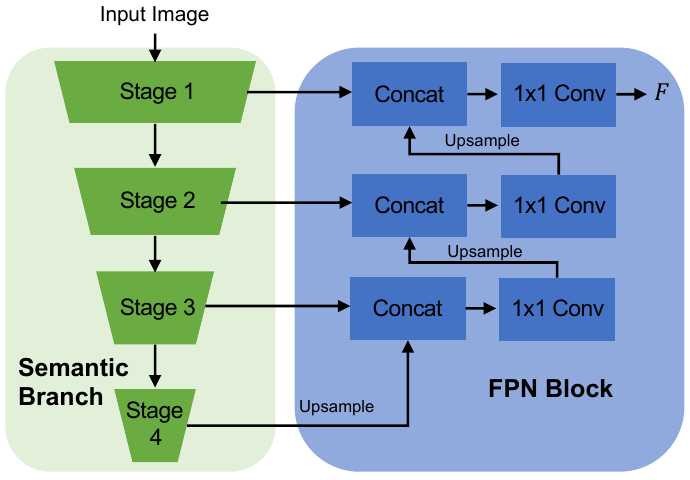}
    \caption{Structure of the FPN block in the proposed Region Selection Gate.}
    \label{fpn_fig}
\end{figure}

\subsection{FPN Block in Region Selection Gate}
The proposed region selection gate employs a FPN block to generate a feature $F$ using the intermediate features of the CNN-based semantic branch, which is then used to assist in selecting informative regions for texture extraction (see Sec. 3.5 of the main paper for details). In Fig. \ref{fpn_fig}, we show the detailed structure of the FPN block. The output channel number of all $1\times 1$ convolution layers in the block is 128. This block integrates multi-level information from different layers. As a result, the generated $F$ contains comprehensive information for the effective key part localization. 

\subsection{Back-Propagation of Improved Semantic Hashing. }In the proposed region selection gate, we employ the improved semantic hashing technique to make the selection operation differentiable. Specifically, for the $k$-th region proposal, a standard Gaussian noise is first added to its score $s^{k}$ to produce $\hat{s}^{k}$. Then two vectors are generated from $\hat{s}^{k}$, including a binary discrete feature $d^{k}$ and a continuously differentiable vector $c^{k}$ (see Eq. 11 of the main paper for details). In forward-propagation, $d^{k}$ is used to make region selection decisions. In back-propagation, we consider the gradient of $c^{k}$ with respect to $\hat{s}^{k}$ an approximation of the gradients for updating the parameters from the discrete gate $d^{k}$. This gradient replacement operation could be realized by $d^{k} = d^{k} + c^{k} - c^{k}.{\rm detach()}$ in PyTorch. During inference, we skip the Gaussian noise sampling step and directly use the discrete output from its original score as the selection decision, i.e., $\mathds{1}\left(s^{k} > 0\right)$.

\begin{table}[t]
    \centering
    \begin{adjustbox}{width=0.8\columnwidth,center}
    \setlength\tabcolsep{14pt}
    \begin{tabular}{l | c}
    \toprule
    Method & Top-1 (\%)\\
    \midrule
    ResNet50 & 76.1  \\
    \midrule
    Ours (with ResNet50 backbone) & 77.9\\
     \bottomrule
    \end{tabular}
    \end{adjustbox}
    \caption{Validation results on ImageNet.}
    \label{imagenet}
\vspace{-1\baselineskip}
\end{table}
\vspace{-0.5\baselineskip}

\section{More Experimental Results} \label{experiment}
\subsection{Experiments on ImageNet}
In addition to fine-grained recognition datasets, we also validate our method on ImageNet, which is a widely-used dataset for general image classification. The results are presented in Table. \ref{imagenet}. As a baseline, ResNet50 achieves Top-1 accuracy of 76.1\%. By using our method with ResNet50 as the semantic branch, we obtain Top-1 of 77.9\%, which outperforms ResNet50 by 1.8\%. Despite achieving higher accuracy, we observe that our method can bring greater improvement on fine-grained datasets than ImageNet.
This can be explained by the different types of features required for different datasets. Specifically, the visual appearances and semantic meanings of different categories in ImageNet are significantly different, allowing us to classify different classes from a global perspective without explicitly exploiting local details. As a result, the high-level semantic information captured by CNN is already sufficient to distinguish different classes, while local detailed textures captured by our method can be less crucial. In contrast, fine-grained recognition datasets often include categories with very similar visual appearances and high-level semantic meanings (e.g., different bird species). These categories are very similar from the global view, only having subtle differences in some local areas. In this scenario, features from deep CNNs are insufficient for classification due to their lack of local detailed features and high-frequency information, as discussed in the Introduction section of our paper. Texture information extracted from our method can serve as an effective supplement to CNN features, significantly improving fine-grained recognition. 

\begin{figure}[t]
    \centering
    \includegraphics[width=0.75\linewidth]{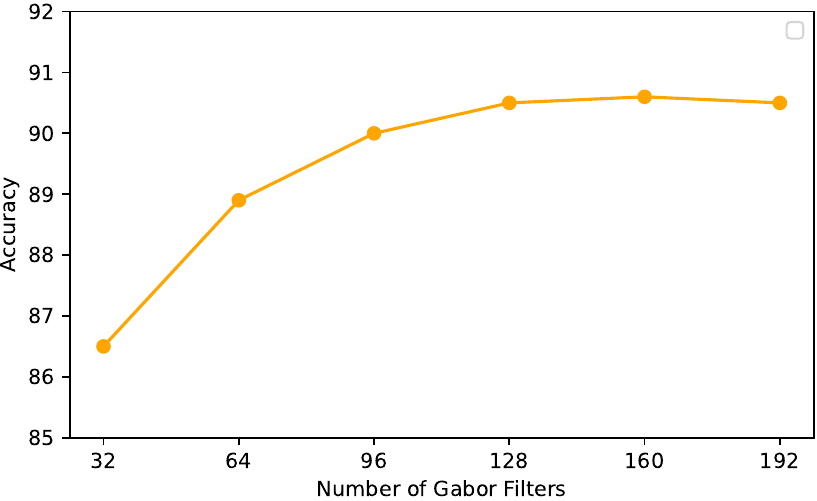}
    \caption{Ablation results of the Gabor filter number.}
    \label{ablation_gabor_num}
\vspace{-0.7\baselineskip}
\end{figure}

\begin{figure}
    \centering
    \includegraphics[width=0.8\linewidth]{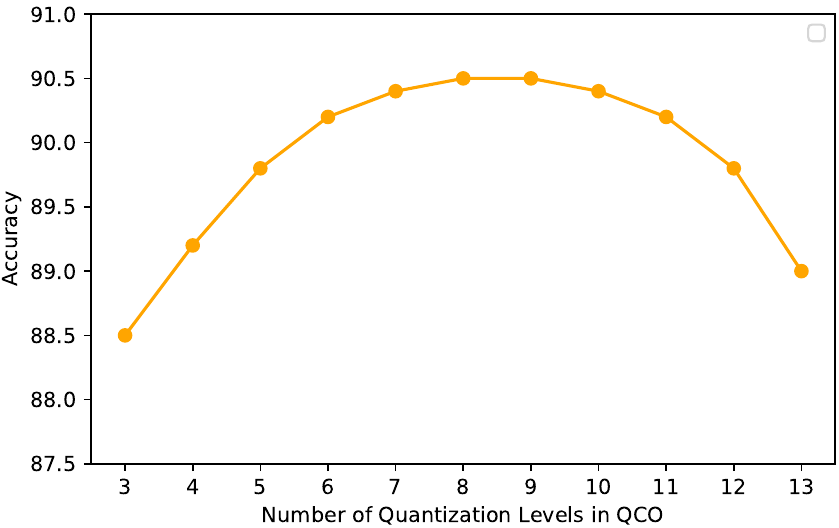}
    \caption{Ablation results for the number of quantization levels in LHO.}
    \label{ablation_quantization_num}
\vspace{-0.7\baselineskip}
\end{figure}

\subsection{Ablation Study of Hyper-Parameters}
In this section, we present ablation results of hyper-parameters used in our method, including the number of Gabor filters, the number of quantization levels in LHO, $\lambda$ in the loss function, and the size that each region is zoomed into. Experiments in this section are conducted on CUB-200-2011 with ResNet50 as the semantic branch. We report the average results of 5 repeated experiments\\

\noindent \textbf{Ablation of Gabor Filter Number}
The texture branch in our approach utilizes $N$ learnable Gabor filters to process input regions. In Fig. \ref{ablation_gabor_num}, we present the validation results of our method using varying numbers of learnable Gabor filters. As shown in Fig. \ref{ablation_gabor_num}, increasing $N$ from 32 to 128 results in an improvement in validation accuracy from 86.5 to 90.5. However, performance improvement becomes insignificant when $N$ exceeds 128. Therefore, we choose $N = 128$ as the optimal number of Gabor filters.\\

\noindent \textbf{Ablation for the Number of Quantization Levels. } The proposed LHO involves a step that quantizes the intensity map into $M$ levels in order to extract statistical information (refer to Eq. 4 and Eq. 5 of the main paper for further details). In Fig. \ref{ablation_quantization_num}, we present the validation results of using different numbers of quantization levels. It is observed that when $M$ is greater than 7 and less than 11, the accuracy remains stable and near 90.5. Conversely, when $M$ is too small, the quantization is coarse, resulting in less effective statistical feature extraction and lower validation accuracy. Furthermore, when $M$ is too large, overfitting may occur to hinder the model's effectiveness. Based on experimental results, we chose 8 as the setting for $M$. It is worth noting that our proposed method consistently outperforms the baseline ResNet50 significantly when $M$ ranges from 3 to 13, thus demonstrating the high effectiveness of our approach.\\

\begin{table}[t]
    \centering
    \begin{adjustbox}{width=0.8\columnwidth,center}
    \setlength\tabcolsep{20pt}
    \begin{tabular}{l | c c}
    \toprule
    $\lambda$ & Accuracy & Flops(G)\\
    \midrule
    0.01 & 90.0 & 25.57\\
    0.05 & 90.2 & 23.26\\
    0.1 & 90.7 & 21.05\\
    0.2 & 90.8 & 20.72\\
    0.3 & 90.5 & 20.65\\
    0.5 & 90.2 & 20.46\\
    1 & 90.0 & 20.20\\
     \bottomrule
    \end{tabular}
    \end{adjustbox}
    \caption{Ablation results of $\lambda$ in Eq. 13 of the main paper. }
    \label{ablation_lamda}
\end{table}

\noindent \textbf{Ablation of $\lambda$ in Loss Function. } As shown in Eq. 13 of the main paper, we use a hyper-parameter $\lambda$ to control the trade-off between the two loss items. In Table. \ref{ablation_lamda}, we present the validation accuracy and average flops when $\lambda$ is set to different values. 

The fluctuation of accuracy is less than 0.8 when $\lambda$ varies from 0.01 to 1, showing that our method is non-sensitive to the choice of hyper-parameter $\lambda$.\\

\begin{figure}
    \centering
    \includegraphics[width=0.8\linewidth]{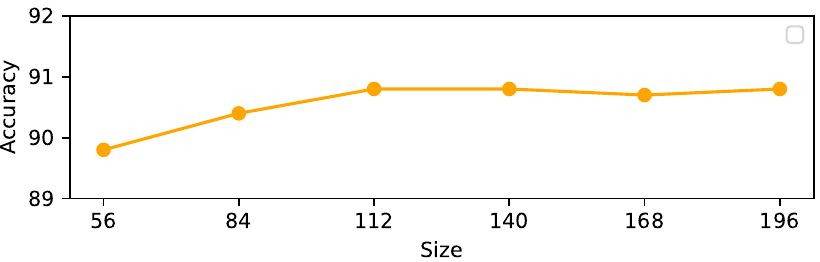}
    \caption{Ablation results for $S$ indicating the size that each region is zoomed into. }
    \label{ablation_size}
\vspace{-0.7\baselineskip}
\end{figure}

\noindent \textbf{Ablation of Size that Each Region is zoomed in. }Each of the selected regions is zoomed into the size $S\times S$ before being feeding into the texture branch for feature extraction. In Fig. \ref{ablation_size}, we show the validation accuracy when setting $S$ to different values. The accuracy keeps stable when $S$ is greater than or equal to 112. Setting a smaller value for $S$ results in a reduced input size for the texture branch and lower computational costs. Thus, we choose 112 to be the setting of $S$.

\begin{table}[t]
    \centering
    \begin{adjustbox}{width=0.9\columnwidth,center}
    \renewcommand{\arraystretch}{1.2}
    \begin{tabular}{c c | c}
    \toprule
    Low-frequency Interval & High-frequency Interval & Accuracy\\
    \midrule
    $[0, \frac{u_{w}}{5}]$ & $[\frac{u_{w}}{5}, u_{w}]$ & 90.1\\
    $[0, \frac{u_{w}}{4}]$ & $[\frac{u_{w}}{4}, u_{w}]$ & 90.5\\
    $[0, \frac{u_{w}}{3}]$ & $[\frac{u_{w}}{3}, u_{w}]$ & 90.5\\
    $[0, \frac{u_{w}}{2}]$ & $[\frac{u_{w}}{2}, u_{w}]$ & 90.8\\
    $[0, \frac{2u_{w}}{3}]$ & $[\frac{2u_{w}}{3}, u_{w}]$ & 90.4\\
    $[0, \frac{3u_{w}}{4}]$ & $[\frac{3u_{w}}{4}, u_{w}]$ & 90.0\\
     $[0, \frac{4u_{w}}{5}]$ & $[\frac{4u_{w}}{5}, u_{w}]$ & 89.4\\  
     \bottomrule
    \end{tabular}
    \end{adjustbox}
    \caption{Ablation results of different divisions for the low-frequency interval and high-frequency interval. $u_{w} = \frac{2\pi S-25}{4\pi S}$} denotes the upper bound of frequency parameter $W$. 
    \label{ablation_division}
\end{table}

\begin{table}[t]
    \centering
    \begin{adjustbox}{width=0.9\columnwidth,center}
    \renewcommand{\arraystretch}{1.4}
    \begin{tabular}{c c | c}
    \toprule
    Low-frequency Filters & High-frequency Filters & Accuracy\\
    \midrule
    $\frac{N}{5}$ & $\frac{4N}{5}$ & 89.8\\
    $\frac{N}{4}$ & $\frac{3N}{4}$ & 90.3\\
    $\frac{N}{3}$ & $\frac{2N}{3}$ & 90.8\\
    $\frac{N}{2}$ & $\frac{N}{2}$ & 90.8\\
    $\frac{2N}{3}$ & $\frac{N}{3}$ & 90.7\\
    $\frac{3N}{4}$ & $\frac{N}{4}$ & 90.0\\
    $\frac{4N}{5}$ & $\frac{N}{5}$ & 89.4\\
     \bottomrule
    \end{tabular}
    \end{adjustbox}
    \caption{Ablation results of different amount allocations for the low-frequency filters and high-frequency filters.} 
    \label{ablation_amount}
\vspace{-1\baselineskip}
\end{table}

\subsection{Ablation of High Frequency Enhancement Strategies. } To alleviate the frequency-bias problem and enhance the high-frequency texture extraction capability of Gabor filters, we propose a high frequency enhancement strategy by setting two value intervals for frequency parameter $W$: $[0, \frac{2\pi S-25}{8\pi S}]$ and $[\frac{2\pi S-25}{8\pi S}, \frac{2\pi S-25}{4\pi S}]$, which are equally divided from the valid range $[0, \frac{2\pi S-25}{4\pi S}]$ of $W$. We then constrain $W$ of $N/2$ Gabor filters to fall between $[0, \frac{2\pi S-25}{8\pi S}]$ and the other $N/2$ filters to fall between $[\frac{2\pi S-25}{8\pi S}, \frac{2\pi S-25}{4\pi S}]$, such that they serve as the low-frequency expert and high-frequency expert respectively. In Table.  \ref{ablation_division}, we present the results obtained from applying different strategies with varying divisions for low-frequency and high-frequency intervals. We denote the upper bound of $W$ as $u_{w}$, which equals $\frac{2\pi S-25}{4\pi S}$. In Table. \ref{ablation_amount}, we present the results of varying amounts of allocations for low-frequency and high-frequency filters. From the results in both tables, it can be observed that the performance remains stable when the division ranges from $\frac{u_{w}}{4}$ to $\frac{2u_{w}}{3}$ and the amount of low-frequency filters ranges from $\frac{N}{3}$ to $\frac{2N}{3}$. The results demonstrate that our proposed method is not sensitive to these specific settings.

\begin{figure}[t]
    \centering
    \begin{minipage}{0.45\linewidth}
    \centering
    \includegraphics[width=1\linewidth]{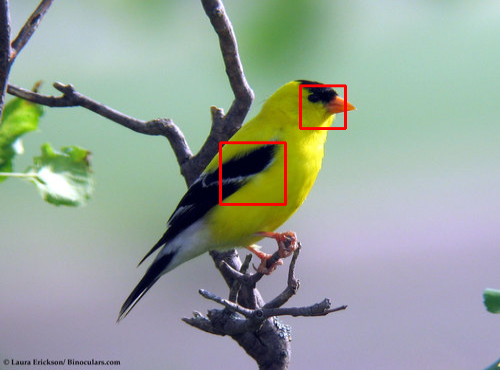} 
    \end{minipage}
    \begin{minipage}{0.45\linewidth}
    \centering
    \includegraphics[width=1\linewidth]{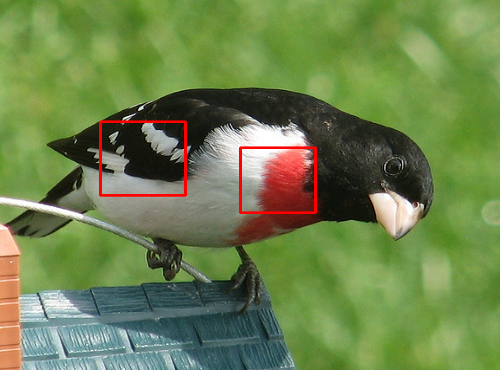}
    \end{minipage}%

    \begin{minipage}{0.45\linewidth}
    \centering
    \includegraphics[width=1\linewidth]{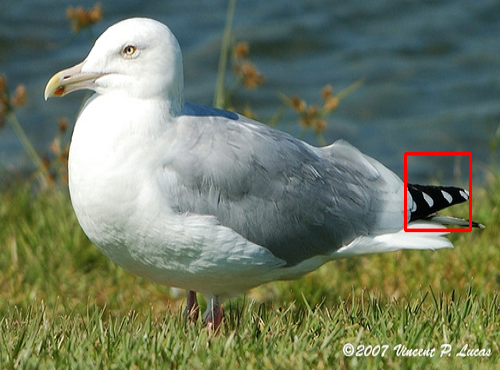} 
    \end{minipage}
    \begin{minipage}{0.45\linewidth}
    \centering
    \includegraphics[width=1\linewidth]{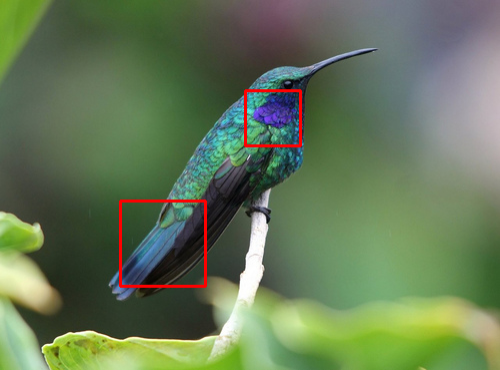}
    \end{minipage}%

    \begin{minipage}{0.45\linewidth}
    \centering
    \includegraphics[width=1\linewidth]{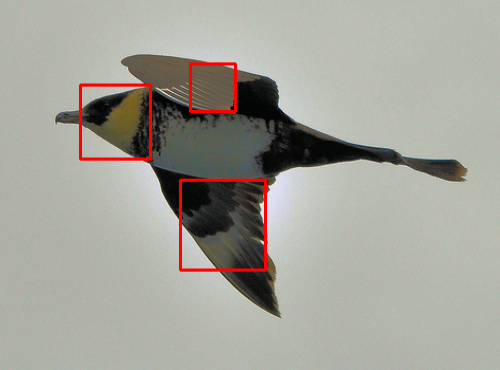} 
    \end{minipage}
    \begin{minipage}{0.45\linewidth}
    \centering
    \includegraphics[width=1\linewidth]{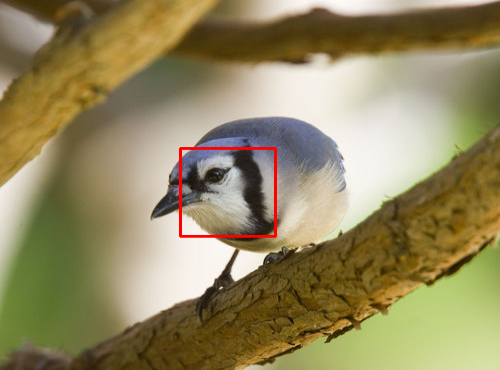}
    \end{minipage}%

    \begin{minipage}{0.45\linewidth}
    \centering
    \includegraphics[width=1\linewidth]{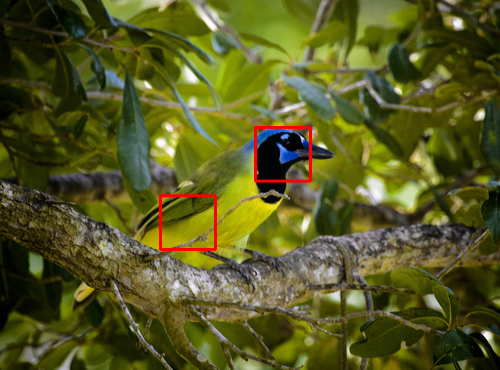} 
    \end{minipage}
    \begin{minipage}{0.45\linewidth}
    \centering
    \includegraphics[width=1\linewidth]{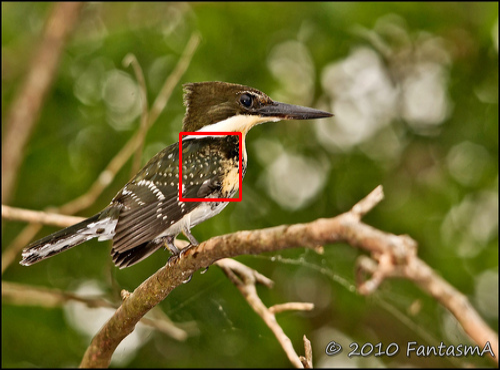}
    \end{minipage}%
    
    \caption{Visualization of selected regions for texture extraction. The selected regions are marked by the red bounding boxes.}
    \label{visual_region}
\end{figure}

\section{More Visualization Results} \label{visual}
\subsection{Visualization of Selected Regions}
In Fig. \ref{visual_region}, we present some visualization results of the selected regions for texture extraction. These regions are marked by the red bounding boxes. The selected regions contain informative texture features that are difficult to be extracted by the vanilla CNNs. Using the proposed texture branch, we can extract effective texture features from these regions to facilitate fine-grained recognition.

\subsection{Visualization for the Output of Gabor Filters}
In Fig. 4 of the main paper, we have provided some visualization results of the outputs obtained from learned Gabor filters. In Fig. \ref{visual_gabor}, we present more visualization results. More specifically, Fig. \ref{visual_gabor} (c) and Fig. \ref{visual_gabor} (d) show the average output of all high-frequency and low-frequency Gabor filters, respectively. As can be observed, the high-frequency filters primarily capture information of undulating areas such as speckles and ripples, whereas the low-frequency filters primarily capture information related to smooth changing areas. Both kinds of information are critical for recognition. By exploiting sufficient and balanced multi-frequency features through the carefully-designed learnable Gabor filters, our method can leverage comprehensive information for effective fine-grained recognition.

\begin{figure}
    \centering
    \includegraphics[width=0.6\linewidth]{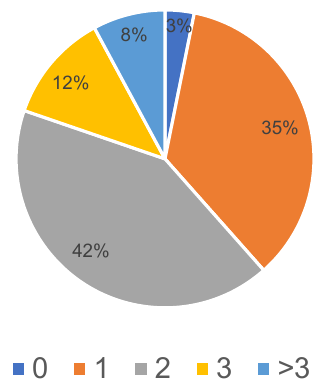}
    \caption{Percentage of images with different numbers of selected regions.}
    \label{selected_num}
\vspace{-0.7\baselineskip}
\end{figure}

\section{Statistical Analysis for the Number of Selected Regions.} \label{analysis} 
\begin{figure}[t]
    \centering
    \begin{minipage}{0.45\linewidth}
    \centering
    \includegraphics[width=1\linewidth]{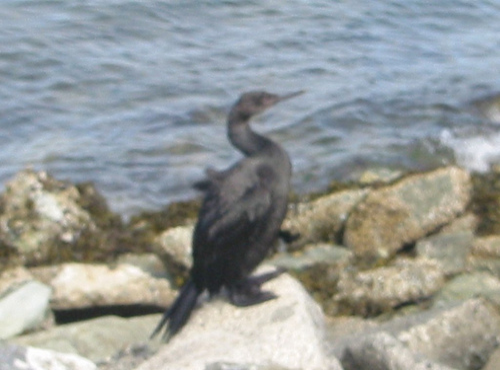} 
    \end{minipage}
    \begin{minipage}{0.45\linewidth}
    \centering
    \includegraphics[width=1\linewidth]{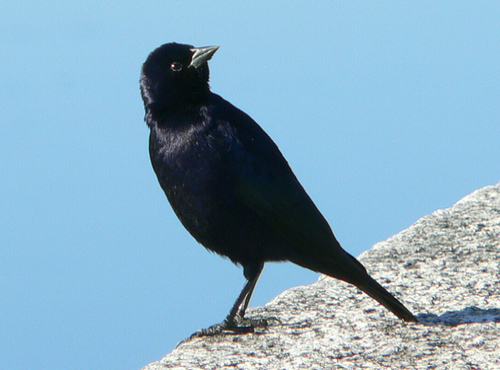}
    \end{minipage}%
    \caption{Two image examples that have no region to be selected for texture extraction. }
    \label{no_region}
\end{figure}

\begin{figure*}[t]
    \centering
    \subfigure[ Image]{
    \begin{minipage}[t]{0.3\linewidth}
    \centering
    \includegraphics[width=0.85\linewidth]{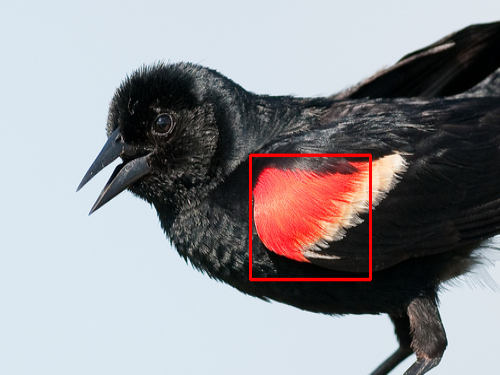}\\
    \hspace{10mm}
    \includegraphics[width=0.85\linewidth]{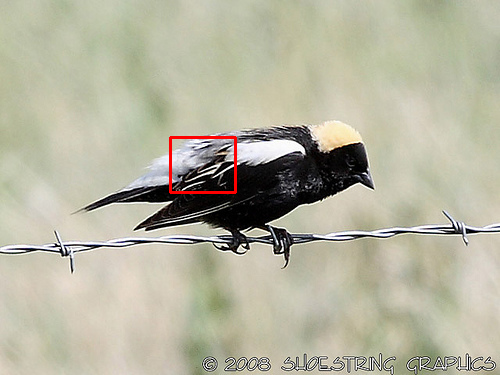}\\
    \hspace{10mm}
    \includegraphics[width=0.85\linewidth]{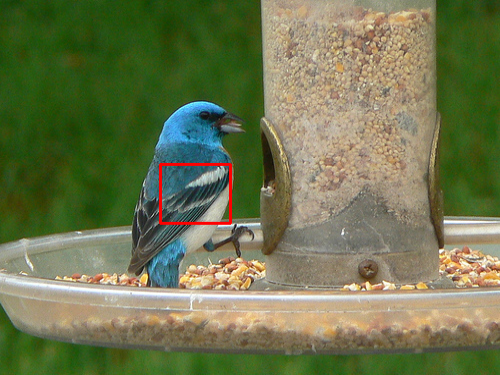}\\
    \hspace{10mm}
    \includegraphics[width=0.85\linewidth]{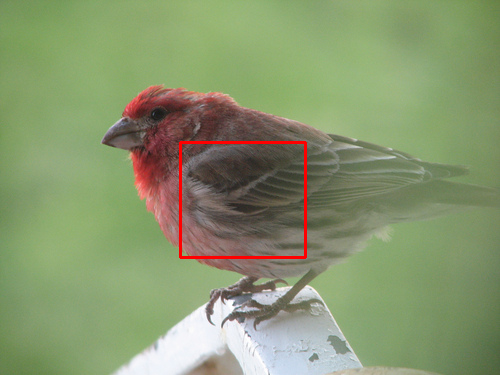}
    \end{minipage}%
    }%
    \subfigure[{ Region}]{
    \begin{minipage}[t]{0.2\linewidth}
    \centering
     \includegraphics[width=0.95\linewidth]{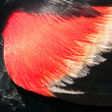}\\
    \hspace{10mm}
    \includegraphics[width=0.95\linewidth]{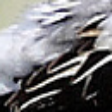}\\
    \hspace{10mm}
    \includegraphics[width=0.95\linewidth]{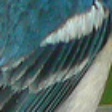}\\
    \hspace{10mm}
    \includegraphics[width=0.95\linewidth]{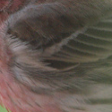}
    \end{minipage}%
    }%
    \subfigure[High-frequency Filters]{
    \begin{minipage}[t]{0.2\linewidth}
    \centering
     \includegraphics[width=0.95\linewidth]{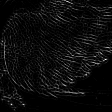}\\
    \hspace{10mm}
    \includegraphics[width=0.95\linewidth]{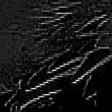}\\
    \hspace{10mm}
    \includegraphics[width=0.95\linewidth]{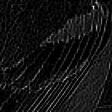}\\
    \hspace{10mm}
    \includegraphics[width=0.95\linewidth]{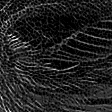}
    \end{minipage}%
    }%
    \subfigure[Low-frequency Filters]{
    \begin{minipage}[t]{0.2\linewidth}
    \centering
     \includegraphics[width=0.95\linewidth]{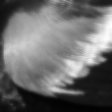}\\
    \hspace{10mm}
    \includegraphics[width=0.95\linewidth]{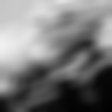}\\
    \hspace{10mm}
    \includegraphics[width=0.95\linewidth]{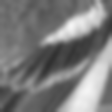}\\
    \hspace{10mm}
    \includegraphics[width=0.95\linewidth]{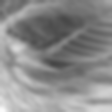}
    \end{minipage}%
    }%
    \caption{Visualization of output from Gabor filters. (a), (b), (c) and (d) present the original images, the selected regions, average output of all high-frequency and low-frequency Gabor filters, respectively.}
    \label{visual_gabor}
\end{figure*}

Fig. \ref{selected_num} displays the percentage of images with various numbers of selected regions for texture extraction. The results indicate that, in general, only a few regions are selected for most images. This minimizes information redundancy and reduces computation costs. Specifically, 35\% of all images have only one region selected, while 42\% of all images have two regions selected for feature extraction. It is worth noting that a very small percentage of images have no regions selected for texture extraction. Fig. \ref{no_region} illustrates two examples of such images. Typically, these images do not contain significant texture information that can facilitate recognition due to the low image quality or the category properties. Therefore, no region is selected for additional texture extraction.

\end{document}